\documentclass[conference]{IEEEtran}
\IEEEoverridecommandlockouts
\usepackage{cite}
\usepackage{threeparttable}
\usepackage{amsmath,amssymb,amsfonts}
\usepackage{ulem} 
\usepackage{algorithmic}
\usepackage{algorithm}
\usepackage{graphicx}
\usepackage{textcomp}
\usepackage{xcolor}
\usepackage{booktabs}
\usepackage{ulem}
\usepackage{amsthm}
\usepackage{multirow} 
\usepackage{caption}
\usepackage{subcaption}

\newtheorem{theorem}{\indent Theorem}
\newtheorem*{prf}{\indent Proof}

\def\BibTeX{{\rm B\kern-.05em{\sc i\kern-.025em b}\kern-.08em
    T\kern-.1667em\lower.7ex\hbox{E}\kern-.125emX}}

\begin{document}

\title{Robust Categorical Data Clustering Guided by Multi-Granular Competitive Learning}

\author{
    \IEEEauthorblockN{Shenghong Cai$^1$, Yiqun Zhang$^1$\textsuperscript{*}, Xiaopeng Luo$^2$, Yiu-Ming Cheung$^3$, Hong Jia$^4$, Peng Liu$^1$}
    \IEEEauthorblockA{$^1$Guangdong University of Technology, Guangzhou, China}
    \IEEEauthorblockA{$^2$Guangzhou Huali College, Guangzhou, China}    
    \IEEEauthorblockA{$^3$Hong Kong Baptist University, Hong Kong SAR, China}
    \IEEEauthorblockA{$^4$Guangdong Provincial Key Laboratory of Intelligent Information Processing, Shenzhen, China}
    \IEEEauthorblockA{3121005074@mail2.gdut.edu.cn, yqzhang@gdut.edu.cn, gordonlok@foxmail.com}
    \IEEEauthorblockA{ymc@comp.hkbu.edu.hk, hongjia1102@szu.edu.cn, liupeng@gdut.edu.cn}
    \IEEEauthorblockA{\textsuperscript{*}Corresponding author}
}

\maketitle

\begin{abstract}
Data set composed of categorical features is very common in big data analysis tasks. Since categorical features are usually with a limited number of qualitative possible values, the nested granular cluster effect is prevalent in the implicit discrete distance space of categorical data. That is, data objects frequently overlap in space or subspace to form small compact clusters, and similar small clusters often form larger clusters. 
However, the distance space cannot be well-defined like the Euclidean distance due to the qualitative categorical data values, which brings great challenges to the cluster analysis of categorical data.
In view of this, we design a Multi-Granular Competitive Penalization Learning (MGCPL) algorithm to allow potential clusters to interactively tune themselves and converge in stages with different numbers of naturally compact clusters.
To leverage MGCPL, we also propose a Cluster Aggregation strategy based on MGCPL Encoding (CAME) to first encode the data objects according to the learned multi-granular distributions, and then perform final clustering on the embeddings.
It turns out that the proposed MGCPL-guided Categorical Data Clustering (MCDC) approach is competent in automatically exploring the nested distribution of multi-granular clusters and highly robust to categorical data sets from various domains. Benefiting from its linear time complexity, MCDC is scalable to large-scale data sets and promising in pre-partitioning data sets or compute nodes for boosting distributed computing.
Extensive experiments with statistical evidence demonstrate its superiority compared to state-of-the-art counterparts on various real public data sets.
\end{abstract}


\section{Introduction}
\begin{figure}[!t]
  \centerline{\includegraphics[width=1\linewidth]{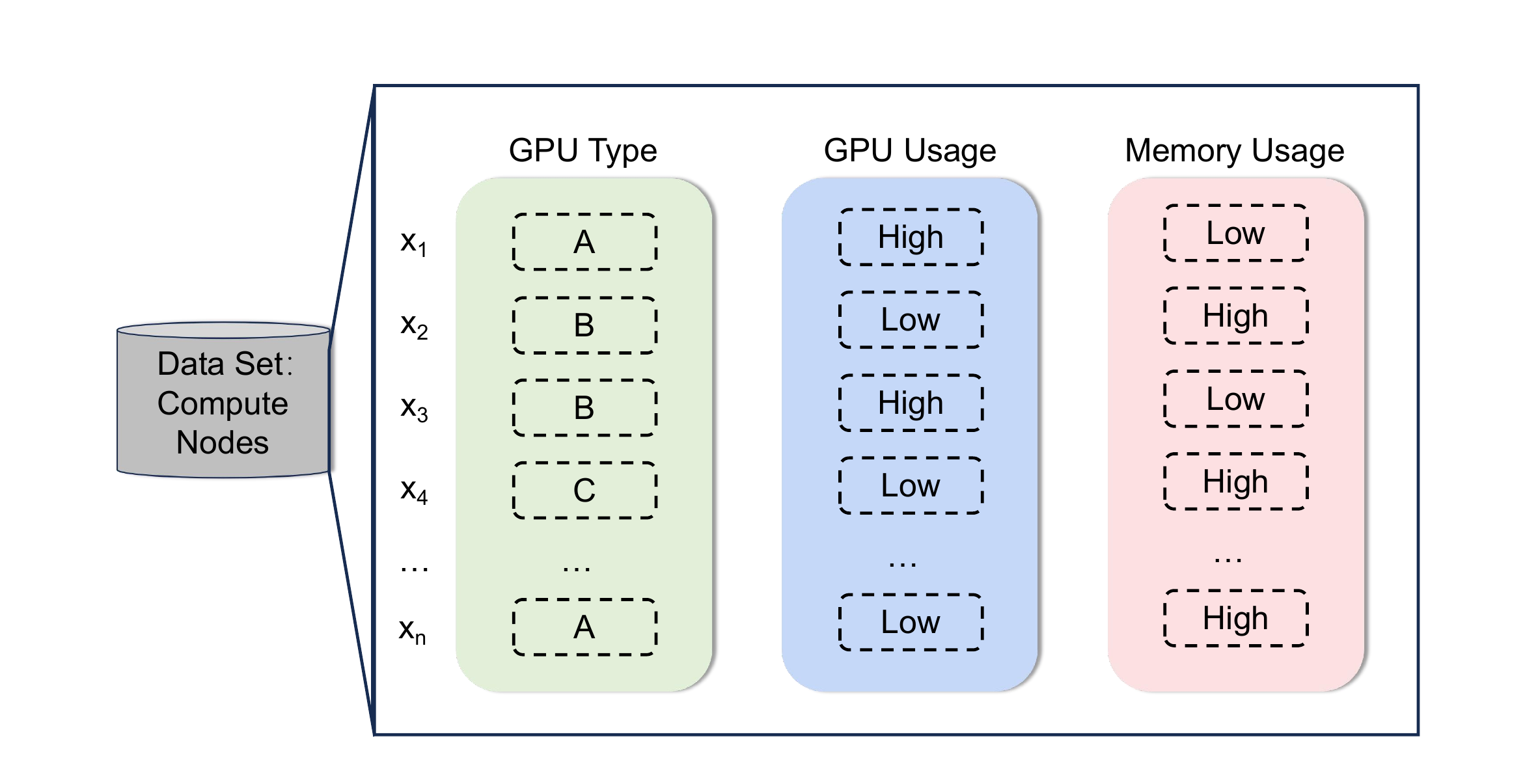}}
  \caption{Three categorical features (i.e., ``GPU Type'', ``GPU Usage'' and ``Memory Usage'') of a data set describing different compute nodes.}
  \label{fig:cd}
\end{figure}

Clustering is one of the most popular unsupervised learning techniques that divides objects into a certain number of clusters where each cluster is usually composed of similar objects \cite{li2022ensemble,xu2015comprehensive}. Clustering can be utilized as a learner for recognition tasks, including anomaly detection \cite{li2021clustering}, recommendation \cite{abbasi2021tourism}, risk detection \cite{caruso2021cluster}, etc. It can also be utilized as a general tool for mining knowledge from data, e.g., latent object distribution \cite{chang2021injury}, potential feature association \cite{song2021fast}, etc. However, most existing clustering is based on numerical data where all the feature values are quantitative and can directly attend arithmetic calculation \cite{yang2022density,chawla2013k,liu2021agglomerative}. Another common type of data, i.e., categorical data \cite{agresti2012categorical,azen2021categorical} composed of qualitative feature values as shown in Fig. \ref{fig:cd}, is usually overlooked.

A common way for categorical data clustering is to encode the qualitative values into quantitative numerical values. However, the encoding process may easily cause information loss \cite{ezugwu2022comprehensive} as the implicit distribution of categorical data is difficult to be appropriately mapped into the Euclidean distance space. By contrast, some conventional categorical data clustering approaches \cite{ezugwu2022comprehensive,huang1997fast,kuo2021metaheuristic,yuan2020dissimilarity}
directly perform clustering by adopting distance metrics that are specifically defined for categorical data. Nevertheless, categorical data set is usually composed of features from various domains, which brings great challenges to defining a universal distance metric. Although hierarchical clustering \cite{murtagh2012algorithms} that outputs a dendrogram reflecting nested data object relationship can be utilized to understand the complex distribution of categorical data, the construction of dendrogram is laborious and may even fail due to the lack of powerful categorical data distance metric. Below we discuss the research progress of the above three streams of methods, i.e., 1) encoding-based, 2) distance defining-based, and 3) hierarchical clustering methods, and then refine the specific cutting-edge problems to be solved in this paper.

For the encoding-based stream, besides the most commonly used one-hot encoding \cite{alamuri2014survey}, more advanced strategies \cite{qian2015space,jian2018cure} that further consider the value-, object-, and feature-level couplings have been proposed for more informative representation. To make the representation learnable with the downstream clustering tasks, representation learning approaches \cite{huang2005automated,zhu2020unsupervised,bai2022categorical} have also been proposed and obtained better categorical data clustering performance. Later, the research \cite{zhang2021learnable} further presents novel learning strategies to circumvent the non-trivial hyper-parameter selection of representation learning. Most recently, \cite{zhang2022het2hom} introduces a multi-view projection technique to extract a more comprehensive representation of categorical feature values. Although the above-mentioned approaches have successively refreshed the clustering performance, they all focus on the improvement of encoding and learning techniques rather than the understanding of complex distributions and cluster effects of categorical data.

For the distance defining-based stream, the most popular Hamming distance \cite{arabie2006studies} assigns distances ``0'' and ``1'' to pairs of identical and different values, respectively, from the perspective of the most basic value matching perspective. Entropy-based measures \cite{barbara2002coolcat,li2004entropy,zhang2019unified,mousavi2023generalized,zhang2020new} and probability-based metrics \cite{le2005association,ahmad2007method,ienco2012context,jia2015new,oskouei2021fkmawcw}, propose to quantify object-cluster affiliation based on more informative data statistics, e.g., occurrence frequency and conditional probability distribution of possible values, and have achieved more satisfactory clustering performance. Noticing the damage to the clustering performance caused by the heterogeneity of numerical and categorical features, some more advanced metrics \cite{cheung2013categorical,jia2017subspace,zhang2022graph} further unify the distance definitions of the two types of features from the perspective of probability. However, they mainly focus on the unification issue, and their performance will degrade when processing pure categorical data. The above measures and metrics are often combined with a partitional clustering algorithm for categorical data clustering. Since most of these algorithms require a given number of sought clusters $k^*$, they are incompetent in exploring and understanding the natural cluster distribution of categorical data.

\begin{figure}[!t]
  \centerline{\includegraphics[width=1\linewidth]{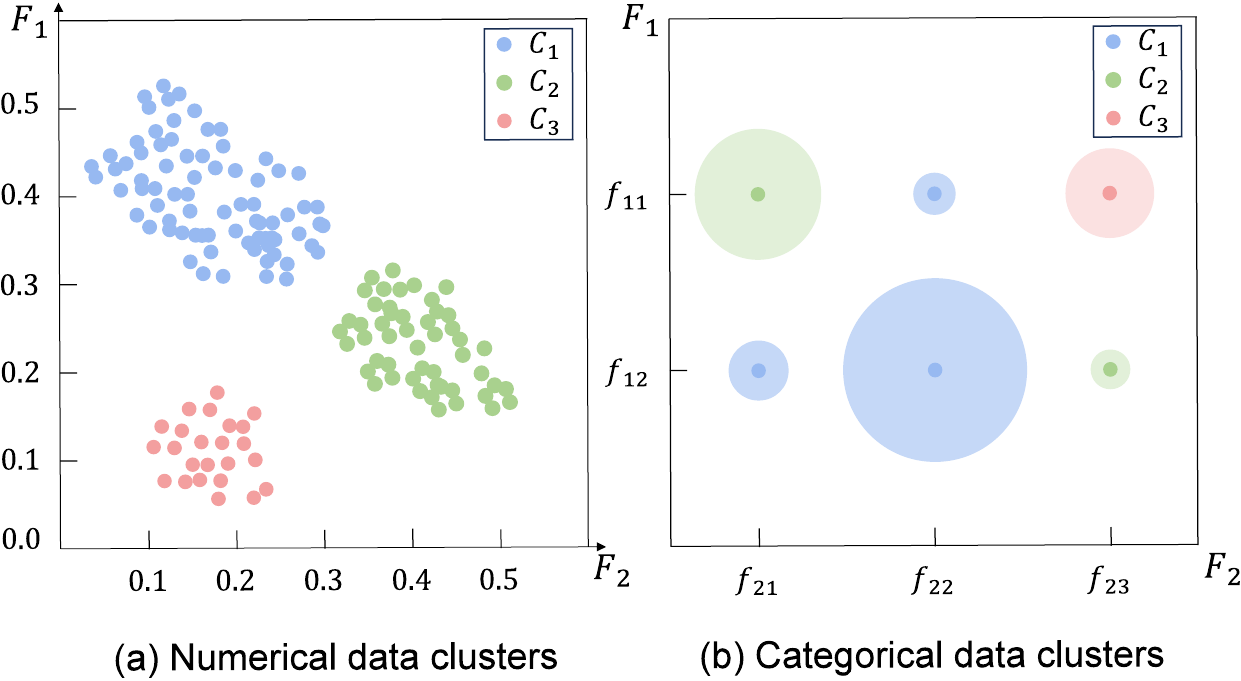}}
  \caption{Comparison of clusters of numerical and categorical data. Since categorical data objects overlap on six points in (b), spheres with different radii are used to indicate the occurrence frequency of overlapping objects. The natural distance structure of categorical data leads to the nested multi-granular cluster effect (e.g., the green cluster is composed of two clusters with different granularity), which brings difficulties to cluster analysis.}
  \label{fig:cnc}
\end{figure}

Hierarchical clustering is considered a promising way for data distribution visualization and understanding, as it produces a tree-like nesting of data objects by recursively linking the current most similar object pairs. Representative link strategies include the conventional average-, complete-, and single-linkage \cite{murtagh2012algorithms}, while recent advanced strategies \cite{jeon2017nc,cheung2018fast,dogan2022k} have also been explored. A link strategy that is specifically designed for categorical data has also been introduced in \cite{guha2000rock}. However, since the adopted dissimilarity metric acts as the basis for linkage computation, and hierarchical clustering lacks a learning mechanism capable of optimizing the metric, metric inappropriateness will all be unavoidably inherited. Moreover, as objects are treated as basic units during the recursive merging, implementing hierarchical clustering to large categorical data will be laborious. Although the most recent work \cite{hu2022significance} attempts to utilize statistical tests to guide the detection of significant clusters, unfortunately, the inherent bias of statistical tests makes them incapable of simultaneously detecting the multi-granular clusters that are prevalent in categorical data as shown in Fig. \ref{fig:cnc}.

It can be seen that the limited feature values of categorical data make the objects overlapped at several points (e.g., the six points in Fig.~\ref{fig:cnc}(b)) in the distance space. The overlapping objects can be viewed as fine-grained clusters. Several such clusters form a larger cluster at a more coarse granularity, and so on, forming the nesting of multi-granular clusters. Based on the aforementioned analysis, it can be concluded that there is still a lack of cluster analysis methods that can effectively reveal the complex nested multi-granular cluster effect and are universally applicable to categorical data sets composed of qualitative features from various conceptual domains.

This paper, therefore, proposes a new cluster analysis framework called MGCPL-guided Categorical Data Clustering (MCDC).
First, a Multi-Granular Competitive Penalization Learning (MGCPL) mechanism is designed to automatically learn object partitions at different granularities by making the learning converge at different numbers of clusters. As MGCPL treats small clusters as the basic unit, the laborious nested relationship analysis is thus greatly alleviated. Compared with hierarchical clustering, the introduced learning mechanism facilitates intelligent multi-granular cluster detection. 
To leverage the analysis results of MGCPL, Cluster Aggregation based on MGCPL Encoding (CAME) has also been proposed to obtain partitional clustering results based on a given number of sought clusters. As its name suggests, clusters explored by MGCPL at each granularity are encoded to obtain informative embeddings, where the multi-granular information may complement each other and thus achieve more accurate clustering. It is worth noting that most existing clustering algorithms can be applied to the embeddings to obtain performance improvements.
It turns out that the proposed method is robust and accurate on real categorical data sets from various domains, and its linear time complexity makes it highly scalable. Extensive experimental evaluations provide sufficient evidence of its effectiveness and efficiency.

The main contributions can be summarized into three-fold:
\begin{itemize}
\item To the best of our knowledge, this is the first attempt to reveal the complex but ubiquitous nested multi-granular cluster effect in categorical data, which is promising in inspiring subsequent works on categorical data analysis. 

\item A new cluster analysis mechanism called MGCPL is proposed to explore the nested multi-granular clusters of categorical data. MGCPL is efficient, and can provide rich data representation information for downstream tasks.

\item An aggregation strategy called CAME is designed to fuse the multi-granular results of MGCPL. CAME achieves more accurate clustering, and its representation can also enhance existing categorical data clustering methods.
\end{itemize}

\section{Preliminaries}

\begin{table}[!t]
\centering
\caption{Frequently used symbols in the paper.}
\label{tbl:1}
\begin{tabular}{c|l}
\toprule
\textbf{Symbols} & \textbf{Explanations}\\
\midrule
\text{$X$} & Data set\\
\text{$F$} & \text{Features}\\
\text{$C$} & \text{Clusters}\\
\text{$n$} & \text{Number of data objects}\\
\text{$d$} & \text{Number of features}\\
\text{$Q$} & \text{Partition matrix of data objets}\\
\text{$C_v$} & \text{The winning cluster}\\
\text{$C_h$} & \text{The rival nearest winner}\\
\text{$\eta$} & \text{Learning rate}\\
\text{$u$} & \text{Cluster weights during competitive learning}\\
$\sigma$ & Number of granularity levels learned by MGCPL\\
\text{$\kappa$} & \text{A series of $k$s learned by MGCPL}\\
\text{$\Gamma$} & \text{Data representation guided by MGCPL}\\
\text{$\Theta$} & \text{Feature weights of representation $\Gamma$}\\
\text{$k^*$} & \text{The true number of clusters}\\
\text{$Z$} & \text{Mode of clusters}\\
\bottomrule
\end{tabular}
\end{table}

This section first briefly introduces basic notations and frequently used symbols (see Table~\ref{tbl:1}), and then presents categorical data distance measurement and competitive learning mechanism that are highly related to the proposed method.

Given a categorical data set $X = \left \{ x_i|i = 1, 2, ..., n\right \}$ with $n$ data objects. Each object $x_i$ is represented by $d$ features $\left \{ F_r|r = 1, 2, ..., d \right \}$. Thus, $x_i$ can be represented as $x_i = \left [ x_{i1}, x_{i2}, ..., x_{id} \right ]^\top$ with $x_{ir} \in dom\left ( F_r \right )$ and $r = 1, 2, ..., d$ where $dom\left ( F_r \right )= \left \{ f_{r1}, f_{r2}, ..., f_{rm_r} \right \}$ contains all the $m_r$ possible values that can be chosen by feature $F_r$. In the common partitional clustering tasks, $X$ should be divided into $k$ clusters $C = \left \{ C_l|l = 1, 2, ..., k \right \}$, i.e., a collection of $k$ disjoint subsets of $X$, where $C_l$ is the set of objects in the $l$th cluster and $X = \bigcup_{l = 1}^{k} C_l$. Since distance measurement plays a key role in most existing categorical data clustering algorithms, we present an object-cluster distance measure for categorical data in the following.

\subsection{Categorical Data Distance Measure}\label{sct:ocd}

To achieve better adaptability between distance definition and clustering task, an object-cluster similarity denoted as $s\left ( x_i, C_l \right )$ can be defined as
\begin{equation}\label{eq:s}
	s\left ( x_i, C_l \right ) = \frac{1}{d} \left[\sum_{r = 1}^{d} s(x_{ir},C_l)\right]
\end{equation}
where 
\begin{equation}\label{eq:sir}
	s(x_{ir}, C_l) = \frac{\Psi _{F_{r} = x_{ir}}(C_l)}{\Psi _{F_{r} \ne NULL}(C_l)}
\end{equation}
is the similarity reflected by the $r$th feature. Note that $\Psi _{F_{r} = x_{ir}}(C_l)$ counts the number of objects in cluster $C_l$ that have value $x_{ir}$ in feature $F_r$, and $\Psi _{F_{r} \ne NULL}(C_l)$ means the number of objects in cluster $C_l$ that have values in the feature $F_r$. Intuitively, $s\left ( x_i, C_l \right )$ is the average of similarities reflected by different features. Then we introduce how the competitive learning mechanism works on categorical data based on the above-defined distance to explore clusters.

\subsection{Competitive Learning Algorithm}\label{sct:cpl} 

Most existing clustering methods assume the known true cluster number $k^*$, which is usually unavailable in real data analysis tasks, especially for data sets with complex distributions like categorical data. Competitive learning \cite{ahalt1990competitive} mechanism is thus designed to learn the true number of clusters. The core idea of competitive learning is to make the initialized clusters compete with each other to eliminate clusters with less importance, and thus its objects will be carved up by the remaining clusters. In this way, by setting a relatively large initial $k$, the algorithm can gradually converge to $k^*$ with a more stable and prominent cluster distribution. Such a learning mechanism is to maximize the overall intra-cluster similarity $S(Q)$:
\begin{equation}\label{eq:Q}
	S(Q) = \sum_{l = 1}^k \sum_{i = 1}^n u_lq_{il}s(x_i,C_l)
\end{equation}
where $q_{il}$ is the $(i, l)$th entry of $Q$, and is computed by
\begin{equation}\label{eq:q}
    q_{il} = \begin{cases}
  & 1,\text{ if } s(x_i,C_l) \ge s(x_i,C_t)\ \  \forall 1\le t\le k \\
  & 0,\ \mathrm {otherwise}.   
  \end{cases}
\end{equation}
$u_l$ is the weight of cluster $C_l$ satisfying $0\le u_l\le 1$ with $l = 1, 2, ..., k$. It measures the importance of $C_j$, and a higher weight indicates that the corresponding cluster is more prominent with less possibility to be eliminated.

During clustering, competitive learning is facilitated as follows. For each input $x_i$, the winning cluster $C_v$ selected from initialized cluster candidates by
\begin{equation}\label{eq:v}
    v = \arg \max_{1\le l\le k} \left [ u_ls(x_i, C_l) \right ] 
\end{equation}
is updated toward $x_i$ by a small step. To avoid the effect that some seed points located in marginal positions will immediately become dead units without learning chance in the subsequent learning process, wining chance of a frequent winning seed point will be gradually reduced. Accordingly, the winning frequency of different clusters can be computed to adjust the selection chance of the winner, and thus Eq.~(\ref{eq:v}) can be re-written as
\begin{equation}\label{eq:new_v}
    v = \arg \max_{1\le l\le k} \left [(1 - \rho_l) u_ls(x_i, C_l) \right ] 
\end{equation}
where 
\begin{equation}\label{eq:rho}
    \rho_l = \frac{g_l}{\sum_{t = 1}^{k} g_t} 
\end{equation}
is a winning ratio computed based on $g_l$, which is the winning times of cluster $C_l$ in the last learning iteration. Accordingly, the weight of cluster $C_v$ is updated by a small step controlled by a small learning rate $\eta$, which can be written as
\begin{equation}\label{eq:award}
    u_l^{new} = u_l^{old} + \eta.   
\end{equation}

Note that the value of initial $k$ should be set at a larger value than $k^*$, i.e., $k\ge k^*$, to ensure that the redundant clusters can be gradually eliminated during the learning process.

\begin{figure*}[!t]
  \centerline{\includegraphics[width=1\linewidth]{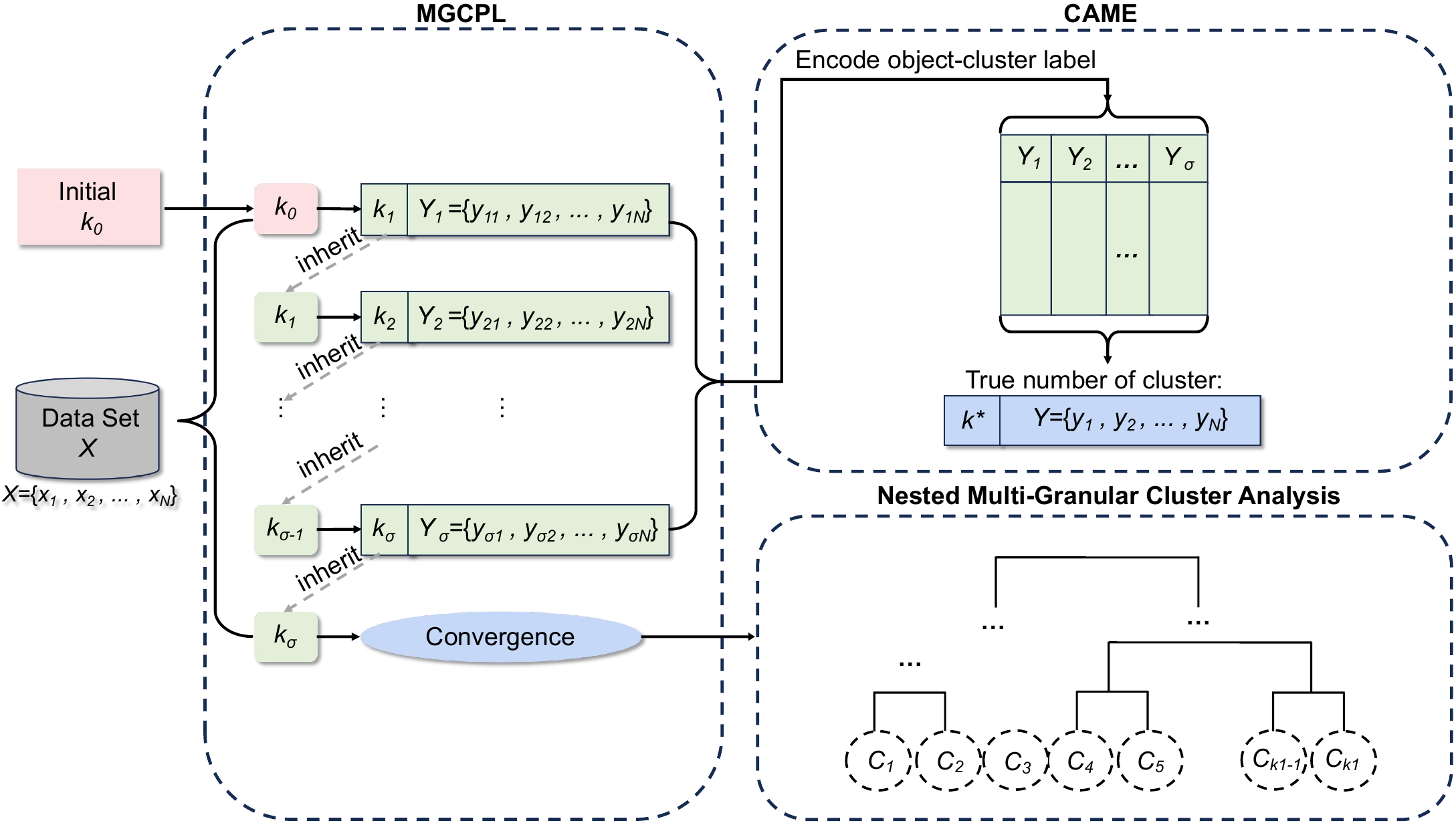}}
  \caption{Pipeline of the proposed method. MGCPL starts its learning with a relatively large initial $k_0$. The initialized clusters compete with each other to eliminate less important ones and obtain $k_1$. By inheriting $k_1$ as the initialization, the learning is re-lunched by clearing the parameters that guide the convergence. Such a process is recursively implemented until converges at $k_\sigma$ where the $k_\sigma$ prominent coarse-grained clusters cannot be further eliminated. The multi-granular results can be utilized for nested cluster distribution analysis, and can also be aggregated by CAME to accurately partition $X$ into $k$ clusters.}
  \label{fig:AGVN}
\end{figure*}

\section{Proposed Method}

This section first presents the MGCPL algorithm for exploring nested multi-granular distribution of clusters, and then introduces the CAME aggregation strategy to combine the multi-granular information provided by MGCPL to obtain the clustering results. The overall pipeline of the cluster analysis method composed of MGCPL and CAME is demonstrated in Fig. \ref{fig:AGVN}. Time complexity analysis, discussions on convergence and distributed computing issues, are provided at the end of this section.

\subsection{MGCPL: Multi-granular Competitive Penalization Learning}

In most real data cluster analysis tasks, it is not the case that a true number of cluster $k^*$ can be known in advance, especially for categorical data with complex non-Euclidean distance space that are difficult to intuitively understand. Therefore, one of the most important issues in clustering is to estimate the most appropriate number of clusters. However, it is common that there are several $k$s suitable for the same data set, as the clusters can exist at different granularities, which is called the multi-granular effect. Such an effect is particularly evident in categorical data, because the categorical features are with limited number of possible values, making data objects overlap in the distance space as discussed in the Introduction.

Hence, for a categorical data set, it is necessary to explore a series of suitable numbers of clusters $\kappa = \left \{ k_1, k_2, ..., k_\sigma  \right \} $ where $k_\sigma$ is the one corresponding to the partition of data objects with the most coarse granularity. As existing competitive learning algorithms aim to find $k^*$ only, we proposed MGCPL to find all the suitable $k$s in $\kappa$. The basic idea is to start the competitive learning with a relatively large initial $k_0$, and let the $k_0$ clusters compete with each other to eliminate less important ones to obtain $k_1$. By inheriting the previously learned $k_1$ as the initialization, the learning is re-lunched by clearing the parameters that guide the convergence. Such a process is recursively implemented until the overall MGCPL converges at $k_\sigma$ where the coarse-grained clusters are prominent enough and no more clusters can be learned to eliminate.

The competitive learning mechanism described by Eq.~(\ref{eq:Q})-Eq.~(\ref{eq:new_v}) only awards the winning cluster while neglecting the rival cluster, which makes the winners gradually absorb the surrounding seed points and thus not conducive to exploring multi-granular clusters. To avoid this, a rival penalization mechanism is introduced. Specifically, for each input $x_i$, the winning cluster $C_v$ selected from the initialized cluster candidates is updated toward $x_i$, while the rival nearest $C_h$ to the winner $C_v$ is determined by
\begin{equation}\label{eq:r}
    h = \arg \max_{1\le l\le k, l \ne v}\left [ (1 - \rho_l) u_ls\left ( x_i, C_l \right )  \right ]
\end{equation}
where $\rho_l$ is defined in Eq.~(\ref{eq:rho}). For each data object $x_i$, when the winning cluster and its rival nearest are determined, $x_i$ will be assigned to the winning cluster $C_v$, and the corresponding winning time is updated by
\begin{equation}\label{eq:win_time}
    g_v = g_v + 1.
\end{equation}
In Eq.~(\ref{eq:r}), $u_l$ is the weight of $C_l$ computed by
\begin{equation}\label{eq:u}
    u_l = \frac{1}{1+e^{\left ( -10\delta_l + 5 \right ) } }
\end{equation}
with $l\in\{1, 2, ..., k\}$. Such a commonly used Sigmoid function form is to ensure a more sensitive updating of rival weights and make its values in the interval [0,1]. Accordingly, the updating of $u_l$ can be accomplished by changing the value of $\delta_l$ instead. Subsequently, the winner $C_v$ is awarded with
\begin{equation}\label{eq:award_new}
    \delta  _v^{new} = \delta  _v^{old} + \eta
\end{equation}
and the rival $C_h$ is penalized with
\begin{equation}\label{eq:penalized}
    \delta  _h^{new} = \delta  _h^{old} - \eta s(x_i, C_l)
\end{equation}
where $\eta$ is a small learning rate. As a result, the rival is penalized a step away from the winner, and thus the rivals obtain more opportunities to explore the cluster distributions in the distance space.

It is usually assumed that all the categorical features have the same contribution during the object-cluster similarity measurement. But in practice, as features are of different importance in forming clusters of different data sets, we improve Eq.~(\ref{eq:s}) with a weighting mechanism by 
\begin{equation}\label{eq:new_sw}
	s\left ( x_i, C_l \right ) = \frac{1}{d} \left[\sum_{r = 1}^{d} \omega_{rl} s(x_{ir},C_l)\right]
\end{equation}
where $\omega_{rl}$ with $0 \le \omega _{rl} \le 1$ is the weight of the $r$th feature to cluster $C_l$, and we have $\sum_{r = 1}^{d} \omega_{rl} = 1$ with $l\in\{1,2, ..., k\}$. Since feature-weight $\omega_{rl}$ is changing with the change of feature-cluster contribution, we use $H_{rl}$ to indicate the contribution of feature $F_r$ to cluster $C_l$. To compute $H_{rl}$, we should first introduce two important terms, i.e., inter-cluster difference $\alpha _{rl}$ and intra-cluster similarity $\beta _{rl}$, where $\alpha _{rl}$ measures the ability of feature $F_r$ in distinguishing cluster $C_l$ from the others, while $\beta _{rl}$ evaluates whether the cluster $C_l$ along the feature $F_r$ has a compact structure. We formulate $\alpha _{rl}$ by
\begin{equation}\label{eq:alpha}
\alpha _{rl} =\frac{1}{\sqrt{2}}\sqrt{\sum_{t = 1}^{m_r}\left ( \frac{\Psi_{F_r = f_{rt}}(C_l)}{\Psi_{F_r \ne NULL}(C_l)} - \frac{\Psi_{F_r = f_{rt}}(X \setminus  C_l)}{\Psi_{F_r \ne NULL}(X \setminus C_l)}\right )^2 }
\end{equation}
and calculate $\beta _{rl}$ by
\begin{equation}\label{eq:beta}
	\beta _{rl} = \frac{1}{n_l}\sum_{x_i \in C_l}\frac{\Psi_{F_r = x_{ir}}(C_l)}{\Psi_{F_r \ne NULL}(C_l)}
\end{equation}
where $n_l$ is the number of objects in $C_l$. When both $\alpha _{rl}$ and $\beta _{rl}$ reach large values, it implies the important contribution of feature $F_r$ in detecting $C_l$, and thus $H_{rl}$ can be obtained by 
\begin{equation}\label{eq:h}
    H_{rl} = \alpha _{rl}\beta _{rl}
\end{equation}
accordingly. Then the corresponding probabilistic feature weight $\omega _{rl}$ can be calculated by
\begin{equation}\label{eq:omega}
	\omega _{rl} = \frac{H_{rl}}{\sum_{t = 1}^{d} H_{tl}}
\end{equation}
with $r\in\{1, 2, ..., d\}$ and $l\in\{1, 2, ..., k\}$.

To facilitate the learning of multi-granular clusters by using the above-described learning process, we initialize a larger number of clusters $k_0$ to launch the learning. When the above-defined competitive penalization learning converges with $k_1$, i.e., an appropriate number of clusters at a fine granularity corresponding to $k_1$ have been explored, we let the learning mechanism inherit $k_1$ and re-launch the learning to explore coarser-grained clusters. To re-launch the learning in a new epoch, all the previous statistics are reset by $g_l=0$, $u_l=1/d$, and $\delta_l=1$ with $l = \{1, 2, ..., k_{\sigma}\}$. Competitive penalization learning is recursively launched until it obtains the same partition as the previous epoch. In this way, we can obtain a series of numbers of clusters with decreasing values $\kappa=\{k_1, k_2, ..., k_\sigma\}$ where $k_\sigma$ is the obtained $k$ with the smallest value. At the same time, we obtain a series of partitions, i.e., clustering results, which can be represented as a collection of object labels, i.e., $\Gamma  = \left \{ Y_1, Y_2, ..., Y_\sigma \right \}$ with $Y_{\sigma} = \left \{ y_{\sigma 1}, y_{\sigma 2}, ..., y_{\sigma n} \right \}$. The whole MGCPL algorithm is summarized as Algorithm ~\ref{alg:1}.

\begin{algorithm}[!t]
	\caption{MGCPL: Multi-Granular Competitive Penalization Learning Algorithm}
	\begin{algorithmic}[1]
		\label{alg:1}
		\REQUIRE Data set $X$, learning rate $\eta$, initialized $k_0$.
		\ENSURE Multi-granular partitions $\Gamma$  = $\left \{ Y_1, Y_2, ..., Y_\sigma \right \}$ and corresponding numbers of clusters $\kappa$ = $\{k_1, k_2, ..., k_\sigma \}$.
		\STATE Initialize convergence = false, $k^{initial}=k_0$.
		\WHILE{convergence = false}
		\STATE change = true, randomly select $k^{initial}$ objects to represent clusters in $C$.
		\WHILE{change = true}
		\FOR{$i$ = $1$ to $n$}
		\STATE Compute $v$ and $h$ by Eqs.~(\ref{eq:new_v}) and (\ref{eq:r}), update $q_{iv}$ by Eq.~(4), update learning variables by Eqs.~(\ref{eq:rho}) and (\ref{eq:win_time})-(\ref{eq:penalized}).
		\ENDFOR
  	\IF{$Q^{new} = Q^{old}$}
		\STATE change = false.
		\ENDIF
		\STATE Update $\omega_{rl}$ by Eq.~(\ref{eq:alpha})-(\ref{eq:omega}).
		\ENDWHILE
        \STATE Set $k^{initial}$ = $k^{new}$, reset $g_l$ = 0, $u_l$ = 1/d and $\delta_l$ = 1 with $l = \{1, 2, ..., k^{new}\}$.
		\IF{$k^{new} = k^{old}$}
		\STATE convergence = true. 
        \ENDIF
		\ENDWHILE
	\end{algorithmic}
\end{algorithm}

\subsection{CAME: Cluster Aggregation based on MGCPL Encoding}\label{sct:came}

The multi-granular cluster distribution information obtained by MGCPL can be utilized to form an informative representation of categorical data. We thus propose a new encoding strategy that uses the object-cluster affiliation $\Gamma$ as the data representation. Then, we implement categorical data clustering on the new representation. The advantage of $\Gamma$ encoding is that it can make full use of the information provided by each granularity of the data set and convert the heterogeneous information provided by the features from different domains into the object-cluster affiliation learned by MGCPL. Moreover, since the features in $\Gamma$ provide object-cluster affiliations at different granularities, their contributions to the final clustering of CAME are usually different in terms of the sought number of clusters $k$. Therefore, we formulate the cluster aggregation in the form of feature importance learning to minimize the objective function $P(Q, \Theta)$ as follows
\begin{equation}\label{eq:P}
	P(Q, \Theta) = \sum_{l = 1}^{k}\sum_{i = 1}^{n}\sum_{r = 1}^{\sigma} q_{il} \theta_{r} d(x_{ir}, Z_{lr})
\end{equation}
where $q_{il}$ is the $(i, l)$th entry of $Q$ defined as
\begin{equation}\label{eq:new_q}
    q_{il} = \begin{cases}
  1,\text{ if } \sum_{r = 1}^{\sigma}\theta_{r}d(x_{ir},Z_{lr}) \le \sum_{r = 1}^{\sigma}\theta_{r}d(x_{ir},Z_{tr})\\
  \ \ \ \ \ \ \ \ \ \ \ \ \ \ \ \ \ \ \ \ \ \ \ \ \ \ \ \ \ \ \ \ \text{for}\ \forall\ t\in\{1,2,...,k\}\\    
  0,\ \mathrm {otherwise}.   
  \end{cases}
\end{equation}

In Eq.~(\ref{eq:P}), the variable $\Theta=\{\theta_1, \theta_2,...,\theta_{\sigma}\}$ is a set of feature weights to be updated during learning. $Z_l$ represents the mode of $l$th cluster with $l = \{k_1, k_2, ..., k_{\sigma}\}$. $s(x_{ir}, Z_{lr})$ is the Hamming distance between feature value of object $x_{ir}$ and feature value of cluster $Z_{lr}$. Note that we use $Z_{lr}$ here to indicate that this is the cluster mode value from representation $\Gamma$ rather than the original data set $X$. In this subsection, all the data values are from $\Gamma$.

The weight $\theta_r$ that reflects the importance of $F_r$ in $\Gamma$ is updated by
\begin{equation}\label{eq:theta}
	\theta _r = \frac{I_r}{\sum_{t = 1}^{\sigma} I_r} 
\end{equation}
where $I_r$ is the overall intra-cluster similarity contributed by $F_r$, which can be written as
\begin{equation}\label{eq:I}
	I_r = \sum_{l = 1}^{k}\sum_{i = 1}^{n}\sum_{r = 1}^{\sigma}\left[1 - d(x_{ir}, Z_{lr})\right].
\end{equation}
A higher intra-cluster similarity of a feature indicates that this feature contributes more to forming clusters with more similar data objects.

Clustering with the above feature weighting can be treated as an optimization problem to minimize Eq.~(\ref{eq:P}). More specifically, we can iteratively solve the following two minimization problems:
\begin{enumerate}
	\item {Fix object partition $Q = \tilde{Q}$, update feature weights $\tilde{\Theta}$};
	\item {Fix feature weights $\Theta = \tilde{\Theta}$, compute object partition $\tilde{Q}$}.
\end{enumerate}
Such a learning process will converge to a minimal solution in a finite number of iterations, and the final clustering result $Q$ can be obtained. We summarize CAME as Algorithm~\ref{alg:2}.

\begin{algorithm}[!t]
	\caption{CAME: Cluster Aggregation based on MGCPL Encoding}
	\begin{algorithmic}[1]
		\label{alg:2}
            \REQUIRE Data representation $\Gamma$, number of clusters $k$.
		\ENSURE Partition $Q$, features importance $\Theta$.
            \STATE Initialize convergence = false and $\theta_r=1/\sigma$ with $r=\{1,2,...,\sigma\}$. 
            \STATE Compute $\tilde{Q}$ according to Eq.~(\ref{eq:new_q}).
            \WHILE{convergence = false}
            \STATE Set $Q$ = $\tilde{Q}$, compute $\tilde{\Theta}$ by Eqs.~(\ref{eq:theta}) and (\ref{eq:I}).
            \STATE Set $\Theta$ = $\tilde{\Theta}$, compute $\tilde{Q}$ by Eq.~(\ref{eq:new_q}).
            \IF{$Q$ = $\tilde{Q}$}
            \STATE convergence = true.
            \ENDIF
            \ENDWHILE
	\end{algorithmic}
\end{algorithm}

\subsection{Time Complexity Analysis}

\begin{theorem}
The time complexity of MGCPL is $O\left(dnk_0\right)$.
\end{theorem}
\begin{prf}
To analyze the complexity in the worst case, we adopt $k_0$ as the initial $k$, $I$ is the maximum number of iterations to make the competitive penalization learning converge. During the object-cluster similarity computation, as $n\times k_0$ pairs of distances should be computed on $d$ features, the time complexity is thus $O(Idnk_0)$ for similarity computation. Similarly, there are $d\times k_0$ weights in total that should be updated based on the statistics obtained by going through all the $n$ data objects, and thus the time complexity for weights updating is $O(dnk_0)$. As for the updating of $g_l$, $u_l$, and $\delta_l$, their time complexity can be omitted compared to that of similarity computation and weights updating. 
Since the above parts will be implemented by $\sigma$ times in Algorithm~\ref{alg:1}, the overall time complexity of MGCPL is $O(\sigma Idnk_0)$. As $\sigma$ and $I$ are both much smaller than $n$, $d$, and $k_0$ in practice, the overall time complexity of MGCPL is thus $O(dnk_0)$.
\qed
\end{prf}

\begin{theorem}
The time complexity of CAME is $O\left(dnk\right)$.
\end{theorem}
\begin{prf}
Assume that the clustering process of CAME needs $T$ iterations to converge. In each iteration, weights of $\sigma$ features are updated by considering the $n$ data objects in $k$ clusters, and thus the time complexity of feature weighing is $O(dnk)$. In each iteration, all the $n$ data objects should also be partitioned into $k$ clusters by computing the object-cluster distances reflected by $\sigma$ features, and thus the time complexity is also $O(dnk)$. For $T$ iterations in total, the overall time complexity is $O(Tdnk)$. Since $T$ can be viewed as a small constant in most cases, the overall time complexity of CAME is thus $O(dnk)$. 
\qed
\end{prf}

\subsection{Discussions on Convergence and Distributed Computing}

The proposed whole clustering approach MCDC is composed of MGCPL in Algorithm~\ref{alg:1} and CAME in Algorithm~\ref{alg:2}. The MGCPL component can be viewed as repeatedly implementing competitive penalization learning \cite{jia2017subspace}, which is a strict gradient decent process that is guaranteed to converge. For the CAME component, it is actually a process of features weighted $k$-modes clustering, which has also been proven to converge in \cite{huang2005automated}. Although we adopt an approximation to more intuitively update the weights by Eq.~(\ref{eq:theta}), such an update strategy is still consistent with the minimization of the objective function in Eq.~(\ref{eq:P}), as features that contribute less on the minimization of the objective function are assigned with smaller importance in the next iteration. Therefore, MCDC well converges on all the data sets we used for the experiments. If strict convergence is required in some scenarios, the weights updating mechanism described by Eqs.~(\ref{eq:theta}) and~(\ref{eq:I}) can be simply replaced by the updating strategy described in \cite{huang2005automated} derived via Lagrange multiplier.

The potential contributions of the proposed multi-granular clustering algorithm to distributed computing systems are mainly two-fold:
\begin{enumerate}
    \item It can be utilized to pre-partition data points into compact subsets to more reasonably allocate them to distributed computing nodes. Specifically, data points described by categorical features are automatically divided into relatively independent and compact micro-clusters, which are automatically merged into larger-scale clusters of different granularities. The multi-granular information obtained in this process can well guide the central server to allocate data sample subsets of different granularities to suitable nodes, flexibly realizing parallel computing without causing significant loss of local correlation information of the data objects.
    \item It can be utilized to pre-divide compute nodes described as the data set shown in Fig. \ref{fig:cd} to form performance-consistent node networks that are more suitable for certain computing tasks. That is, computing nodes are automatically grouped into multi-granular clusters according to their categorical features. The nodes in the same cluster have relatively consistent computing performance and features, and can thus collaborate more efficiently to complete distributed computing tasks. Therefore, the obtained multi-granular computing node clusters can flexibly guide the selection of uniform nodes according to computing task requirements.
\end{enumerate}

\section{Experiment}

This section introduces the experimental design and the selection of counterparts, validity indices, and data sets. Then five parts of experimental results are demonstrated with in-depth discussions for performance evaluation of the proposed MGCPL-guided Categorical Data Clustering (MCDC).

\subsection{Experimental Settings}

\textbf{Five Experiments} are conducted to evaluate the proposed method from different perspectives, which are summarized below.
\begin{itemize}
\item Clustering performance evaluation: The proposed MCDC method is compared with existing representative clustering approaches by quantifying their clustering performance using different mainstream validity indices. 
\item Significance test: Wilcoxon signed rank test is conducted on the performance of the compared approaches. A rejection of the null hypothesis indicates a significant outperforming of our proposed method against the counterparts.
\item Ablation Study: MCDC is ablated into different versions by successively removing its main technical components, and the performance of these versions is compared to illustrate the effectiveness of the MCDC components. 
\item Learning process evaluation: MGCPL will converge to different $k$s during its learning. We visualize the changing of $k$ with the optimal $k^*$ to illustrate the effectiveness of the multi-granular cluster learning mechanism.   
\item Computational efficiency evaluation: MGCPL can be viewed as an efficient alternative to hierarchical clustering. We thus plotting and comparing its execution time under different $n$s, $d$s, and $k$s with the counterparts. 
\end{itemize}

\textbf{Nine Counterparts} are compared in the comparative experiments, including six representative clustering methods, two variants of MCDC that adopt and enhance two existing categorical data clustering algorithms, and MCDC itself. The six representative counterparts are the conventional $k$-modes \cite{huang1997fast} proposed for partitional clustering and ROCK \cite{guha2000rock} for hierarchically clustering categorical data. Four recent advanced clustering methods, i.e., WOCIL \cite{jia2017subspace}, GUDMM \cite{mousavi2023generalized}, FKMAWCW \cite{oskouei2021fkmawcw}, and ADC \cite{zhang2022graph} are also chosen for more convincing comparison. WOCIL is proposed for automatically learning the clusters of mixed data, i.e., data composed of both categorical and numerical features. GUDMM introduces a generalized multi-aspect distance measure based on mutual information. FKMAWCW is a fuzzy k-modes-based approach that learns weights of features to clusters during clustering. ADC utilizes a graph-based dissimilarity measurement for cluster analysis of data composed of any type of features. GUDMM and FKMAWCW are also applied to the multi-granular encoding output of our proposed MCDC. The formed clustering approaches are named MCDC+GUDMM and MCDC+FKMAWCW, respectively. For simplicity, they are abbreviated as MCDC+G. and MCDC+F. hereinafter. Hyper-parameters of the compared methods (if any) are set according to the corresponding source paper. Learning rate $\eta$ and $k_0$ of MCDC is set at $\eta=0.03$ and $k_0=\sqrt{n}$, respectively. For all the compared methods, the sought number of clusters is set at $k^*$ corresponding to each data set as shown in Table~\ref{tbl:2}.

\textbf{Four Validity Indices}\footnote{https://scikit-learn.org/stable/modules/clustering.html\#clustering-evaluation} are utilized to measure the clustering performance. Clustering Accuracy (ACC) is a commonly used index that ranges from 0 to 1. It computes the ratio of the number of correctly clustered objects to the total number of objects. Adjusted Rand Index (ARI) calculates the consistency of obtained clustering results and true labels by comparing their pairwise matching. Its values range from -1 to 1. Adjusted Mutual Information (AMI) is based on mutual information, which quantifies the matching between obtained clustering results and true labels from the perspective of information theory. Its values also range from -1 to 1. The Fowlkes-Mallows (FM) score is defined as the geometric mean of the pairwise precision and recall, and ranges from 0 to 1. All the adopted indices reflect a better clustering performance with a higher value.

\textbf{Ten Data Sets} are utilized to conduct a comprehensive evaluation. Among them, eight data sets are representative ones downloaded from the UCI Machine Learning Repository\footnote{https://archive.ics.uci.edu/}. Two synthetic data sets with large $n$ and $d$ are generated with well-separated clusters for the efficiency evaluation. All the data sets are categorical ones, and data objects with missing values are omitted before conducting experiments. Detailed statistics of the data sets are shown in Table~\ref{tbl:2}.

\begin{table}[!t]
    \centering
    \caption{Statistics of the 8 data sets.$d$, $n$, and $k^*$ indicate the number of features, the number of objects, and the true number of clusters, respectively.}
    \label{tbl:2}
    \begin{tabular}{c|c c|c c c}
    \toprule
    \textbf{No.} & \textbf{Data Set} & \textbf{Abbrev.} & \textbf{$d$} & \textbf{$n$} & \textbf{$k^*$}\\
    \midrule
    \text{1} & \text{Car Evaluation} & \text{Car.} & \text{6} & \text{1728} & \text{4}\\
    \text{2} & \text{Congressional} & \text{Con.} & \text{16} & \text{435} & \text{2}\\
    \text{3} & \text{Chess} & \text{Che.} & \text{36} & \text{3196} & \text{2}\\
    \text{4} & \text{Mushroom} & \text{Mus.} & \text{22} & \text{8124} & \text{2}\\
    \text{5} & \text{Tic Tac Toe} & \text{Tic.} & \text{9} & \text{958} & \text{2}\\
    \text{6} & \text{Vote} & \text{Vot.} & \text{16} & \text{232} & \text{2}\\
    \text{7} & \text{Balance} & \text{Bal.} & \text{4} & \text{625} & \text{3}\\
    \text{8} & \text{Nursery} & \text{Nur.} & \text{8} & \text{12960} & \text{5}\\
    \text{9} & \text{Synthetic (with large $n$)} & \text{Syn\_n} & \text{10} & \text{200000} & \text{3}\\
    \text{10} & \text{Synthetic (with large $d$)} & \text{Syn\_d} & \text{1000} & \text{20000} & \text{3}\\
    \bottomrule
    \end{tabular}
\end{table}

\subsection{Clustering Performance Evaluation}

\begin{table*}[!t]
    \centering
    \caption{Clustering performance w.r.t., ACC, ARI, AMI, and FM on categorical data sets. MCDC+G. and MCDC+F. are the variants of MCDC adopting GUDMM and FKMAWCW, respectively. The best and second-best results on each data set are highlighted using \textbf{boldface} and \underline{underline}, respectively.}
    \label{tbl:3}
    \begin{tabular}{c|c|c c c c c c | c c c}
    \toprule
    \textbf{Index}&\textbf{Data} & \textbf{K-MODES} & \textbf{ROCK} & \textbf{WOCIL} & \textbf{FKMAWCW} & \textbf{GUDMM} & \textbf{ADC} & \textbf{MCDC} & \textbf{MCDC+G.} & \textbf{MCDC+F.}\\
    \midrule
        \multirow{8}{*}{ACC}
        &Car. & 0.372±0.00 & 0.326±0.00 & 0.270±0.00 & 0.371±0.00 & 0.372±0.00 & 0.361±0.00 & \underline{0.373±0.00} & 0.270±0.00 & \textbf{0.414±0.00}  \\
        &Con. & \underline{0.866±0.00} & 0.506±0.00 & \textbf{0.874±0.00} & 0.796±0.01 & 0.818±0.00 & \textbf{0.874±0.00} & \textbf{0.874±0.00} & \textbf{0.874±0.00} & \textbf{0.874±0.00}  \\ 
        &Che. & 0.551±0.00 & 0.505±0.00 & 0.531±0.00 & 0.561±0.00 & 0.554±0.00 & 0.548±0.00 & \underline{0.578±0.00} & 0.547±0.00 & \textbf{0.585±0.00}  \\ 
        &Mus. & 0.740±0.02 & 0.509±0.00 & 0.678±0.00 & 0.000±0.00 & 0.501±0.00 & \underline{0.752±0.02} & 0.710±0.00 & 0.613±0.00 & \textbf{0.784±0.00}  \\ 
        &Tic. & 0.557±0.00 & \textbf{0.674±0.00} & 0.526±0.00 & 0.538±0.00 & 0.507±0.00 & 0.535±0.00 & 0.602±0.00 & 0.642±0.00 & \underline{0.646±0.00}  \\ 
        &Vot. & 0.869±0.00 & 0.500±0.00 & \underline{0.888±0.00} & 0.778±0.01 & 0.828±0.00 & \underline{0.888±0.00} & \textbf{0.905±0.00} & \textbf{0.905±0.00} & \textbf{0.905±0.00}  \\
        &Bal. & 0.448±0.00 & \underline{0.496±0.00} & 0.419±0.00 & 0.463±0.00 & 0.000±0.00 & 0.442±0.00 & 0.464±0.00 & 0.453±0.00 & \textbf{0.506±0.00}  \\ 
        &Nur. & 0.332±0.00 & 0.000±0.00 & 0.239±0.00 & 0.315±0.00 & 0.000±0.00 & 0.337±0.00 & \underline{0.340±0.00} & 0.305±0.00 & \textbf{0.432±0.00}  \\
        \midrule
        \multirow{8}{*}{ARI}&Car. & 0.027±0.00 & 0.023±0.00 & 0.001±0.00 & -0.002±0.00 & \textbf{0.054±0.00} & 0.017±0.00 & \underline{0.051±0.00} & 0.001±0.00 & 0.027±0.00  \\ 
        &Con. & \underline{0.536±0.00} & -0.004±0.00 & \textbf{0.557±0.00} & 0.385±0.05 & 0.394±0.00 & \textbf{0.557±0.00} & \textbf{0.557±0.00} & \textbf{0.557±0.00} & \textbf{0.557±0.00}  \\ 
        &Che. & 0.014±0.00 & -0.001±0.00 & 0.003±0.00 & 0.020±0.00 & 0.012±0.00 & 0.015±0.00 & \underline{0.024±0.00} & 0.008±0.00 & \textbf{0.028±0.00}  \\ 
        &Mus. & 0.303±0.07 & -0.001±0.00 & 0.125±0.00 & 0.000±0.00 & -0.003±0.00 & \underline{0.321±0.06} & 0.186±0.01 & 0.051±0.00 & \textbf{0.323±0.00} \\ 
        &Tic. & 0.017±0.00 & \textbf{0.120±0.00} & 0.000±0.00 & -0.002±0.00 & -0.001±0.00 & 0.007±0.00 & 0.038±0.00 & \underline{0.079±0.00} & 0.062±0.00   \\ 
        &Vot. & 0.543±0.00 & -0.004±0.00 & \underline{0.600±0.00} & 0.349±0.05 & 0.427±0.00 & \underline{0.600±0.00} & \textbf{0.655±0.00} & \textbf{0.655±0.00} & \textbf{0.655±0.00}  \\ 
        &Bal. & 0.027±0.00 & \textbf{0.080±0.00} & 0.005±0.00 & 0.055±0.00 & 0.000±0.00 & 0.025±0.00 & 0.052±0.00 & 0.016±0.00 & \underline{0.079±0.00} \\ 
        &Nur. & 0.049±0.00 & 0.000±0.00 & 0.002±0.00 & 0.028±0.00 & 0.000±0.00 & \underline{0.052±0.00} & 0.051±0.00 & 0.004±0.00 & \textbf{0.166±0.00} \\ 
         \midrule
        \multirow{8}{*}{AMI}&Car. & 0.049±0.00 & 0.050±0.00 & 0.003±0.00 & 0.082±0.00 & \underline{0.117±0.00} & 0.047±0.00 & \textbf{0.123±0.00} & 0.003±0.00 & 0.015±0.00  \\ 
        &Con. & \underline{0.473±0.00} & 0.001±0.00 & \textbf{0.484±0.00} & 0.337±0.03 & 0.380±0.00 & \textbf{0.484±0.00} & \textbf{0.484±0.00} & \textbf{0.484±0.00} & \textbf{0.484±0.00}  \\ 
        &Che. & 0.012±0.00 & 0.000±0.00 & 0.003±0.00 & \textbf{0.021±0.00} & 0.011±0.00 & 0.015±0.00 & \underline{0.020±0.00} & 0.005±0.00 & \underline{0.020±0.00}  \\ 
        &Mus. & \underline{0.280±0.05} & 0.000±0.00 & 0.235±0.00 & 0.000±0.00 & 0.044±0.00 & \textbf{0.347±0.04} & 0.209±0.01 & 0.036±0.00 & 0.248±0.00 \\ 
        &Tic. & 0.012±0.00 & \textbf{0.120±0.00} & 0.007±0.00 & 0.005±0.00 & 0.000±0.00 & 0.006±0.00 & 0.020±0.00 & \underline{0.058±0.00} & 0.023±0.00 \\ 
        &Vot. & 0.457±0.00 & 0.000±0.00 & \underline{0.522±0.00} & 0.301±0.03 & 0.417±0.00 & \underline{0.522±0.00} & \textbf{0.566±0.00} & \textbf{0.566±0.00} & \textbf{0.566±0.00}  \\ 
        &Bal. & 0.026±0.00 & 0.071±0.00 & 0.008±0.00 & 0.048±0.00 & 0.000±0.00 & 0.026±0.00 & \underline{0.083±0.00} & 0.017±0.00 & \textbf{0.089±0.00}  \\ 
        &Nur. & 0.060±0.00 & 0.000±0.00 & 0.004±0.00 & 0.043±0.00 & 0.000±0.00 & 0.061±0.00 & \underline{0.077±0.00} & 0.022±0.00 & \textbf{0.208±0.00}  \\ 
        \midrule
        \multirow{8}{*}{FM}&Car. & 0.409±0.00 & 0.394±0.00 & 0.369±0.00 & 0.406±0.00 & \underline{0.413±0.00} & 0.401±0.00 & 0.407±0.00 & 0.369±0.00 & \textbf{0.434±0.00}   \\ 
        &Con. & \underline{0.774±0.00} & 0.518±0.00 & \textbf{0.784±0.00} & 0.711±0.01 & 0.754±0.00 & \textbf{0.784±0.00} & \textbf{0.784±0.00} & \textbf{0.784±0.00} & \textbf{0.784±0.00}  \\ 
        &Che. & 0.544±0.00 & 0.525±0.00 & 0.507±0.00 & \textbf{0.578±0.00} & 0.554±0.00 & 0.555±0.00 & \underline{0.573±0.00} & 0.519±0.00 & 0.532±0.00 \\ 
        &Mus. & 0.667±0.02 & 0.525±0.00 & 0.657±0.00 & 0.000±0.00 & \underline{0.687±0.00} & \textbf{0.721±0.01} & 0.640±0.00 & 0.544±0.00 & 0.662±0.00  \\ 
        &Tic. & 0.538±0.00 & \underline{0.581±0.00} & 0.527±0.00 & 0.547±0.00 & 0.524±0.00 & 0.526±0.00 & 0.548±0.00 & 0.562±0.00 & \textbf{0.612±0.00} \\ 
        &Vot. & 0.772±0.00 & 0.500±0.00 & \underline{0.800±0.00} & 0.696±0.01 & 0.734±0.00 & \underline{0.800±0.00} & \textbf{0.827±0.00} & \textbf{0.827±0.00} & \textbf{0.827±0.00}  \\ 
        &Bal. & 0.426±0.00 & 0.441±0.00 & 0.437±0.00 & 0.424±0.00 & 0.000±0.00 & 0.425±0.00 & \textbf{0.464±0.00} & \underline{0.460±0.00} & 0.452±0.00  \\ 
        &Nur. & 0.303±0.00 & 0.000±0.00 & 0.260±0.00 & 0.306±0.00 & 0.000±0.00 & 0.305±0.00 & 0.309±0.00 & \underline{0.321±0.00} & \textbf{0.396±0.00} \\ 
    \bottomrule
    \end{tabular}
\end{table*}

The clustering performance of the proposed MCDC is compared with the counterparts on all eight categorical data sets in Table~\ref{tbl:3}. Each result in the table is obtained by executing the corresponding method by 50 times and taking the average performance with standard deviation. Please note that MCDC+G. and MCDC+F. are the two variants of MCDC adopting GUDMM and FKMAWCW as the clustering algorithms, respectively. Comparing their performance with the original GUDMM and FKMAWCW can reflect the effectiveness of the proposed MCDC in enhancing the clustering performance of existing clustering methods. In the table, the best and the second-best results on each data set are highlighted using \textbf{boldface} and \underline{underline}, respectively.

It can be observed from Table~\ref{tbl:3} that MCDC outperforms its counterparts on most data sets. Although MCDC does not perform the best on some data sets, it still ranks second or with a performance that is very close to the best performer. Thus, it can be concluded that, in general, MCDC demonstrates its superiority in terms of both accuracy and robustness. From the table, we can also see that ROCK, FKMAWCW, and GUDMM have unsatisfactory performance on some data sets. This is because they sometimes cannot obtain the pre-set number of clusters and are judged as failed. Moreover, ROCK, WOCIL, and three MCDC variants perform very stable because Rock is a hierarchical clustering approach without random initialization, and WOCIL adopts a very stable initialization mechanism. The performance of MCDC is also very stable because the learned multi-granular information complements each other to form a comprehensive and stable representation of different data sets.

As for MCDC+G. and MCDC+F., it can be seen that the performance of the corresponding GUDMM and FKMAWCW is obviously boosted in most cases by MCDC. This indicates that the proposed MCDC is effective in enhancing different categorical data clustering methods. It can be seen that MCDC, MCDC+G., and MCDC+F. achieve obviously better clustering performance than the other counterparts. Moreover, MCDC+F. performs the best in general. This may be because the corresponding FKMAWCW is a fuzzy clustering algorithm that suits categorical data better. More specifically, fuzzy algorithms can more appropriately describe the object-cluster similarity based on the statistics during clustering, and can thus better exploit the multi-granular information in the embeddings provided by MCDC.

\subsection{Significance Test}

To provide statistical evidence of the superiority of MCDC, we conduct a significance test using the Wilcoxon signed-rank test based on the clustering performance shown in Table \ref{tbl:3}, and demonstrate the test results in Table~\ref{tbl:4}. The best-performing version of MCDC, i.e., MCDC+F., is compared with each of the counterparts at a 90\% confidence interval. We use ``+'' to indicate a rejection of the null hypothesis, which means MCDC+F. significantly outperforms the corresponding counterpart w.r.t. certain validity index.

It can be seen from Table~\ref{tbl:4} that MCDC significantly outperforms its counterparts under almost all the indices, which obviously illustrates the superiority of MCDC. Although the test does not show a significant advantage of MCDC in comparison with K-MODES and ROCK in terms of AMI, MCDC+F. still outperforms them on most data sets as shown in Table~\ref{tbl:3}.

\begin{table}[!t]
    \centering
    \caption{Results of two-tailed Wilcoxon signed-rank test conducted with confidence interval 90\% (i.e. $\alpha$ = 0.1). The symbol ``+'' indicates that MCDC+F. performs significantly better than a certain counterpart, while ``-'' indicates that there is no significant difference between the two methods.}
    \label{tbl:4}
    \begin{tabular}{c|c c c c}
    \toprule
    \textbf{Method} & \textbf{ACC} & \textbf{ARI} & \textbf{AMI}& \textbf{FM} \\
    \midrule
    \text{K-MODES} & \text{+} & \text{+} & \text{-} & \text{+}\\
    \text{ROCK} & \text{+} & \text{+} & \text{-} & \text{+}\\
    \text{WOCIL} & \text{+} & \text{+} & \text{+} & \text{+}\\
    \text{FKMAWCW} & \text{+} & \text{+} & \text{+} & \text{+}\\
    \text{GUDMM} & \text{+} & \text{+} & \text{+} & \text{+}\\
    \text{ADC} & \text{+} & \text{+} & \text{-} & \text{-}\\
    \bottomrule
    \end{tabular}
\end{table}

\subsection{Ablation Study}

\begin{figure}[!t]
	\centering
	\begin{subfigure}{0.49\columnwidth}
		\includegraphics[width=1.1\linewidth]{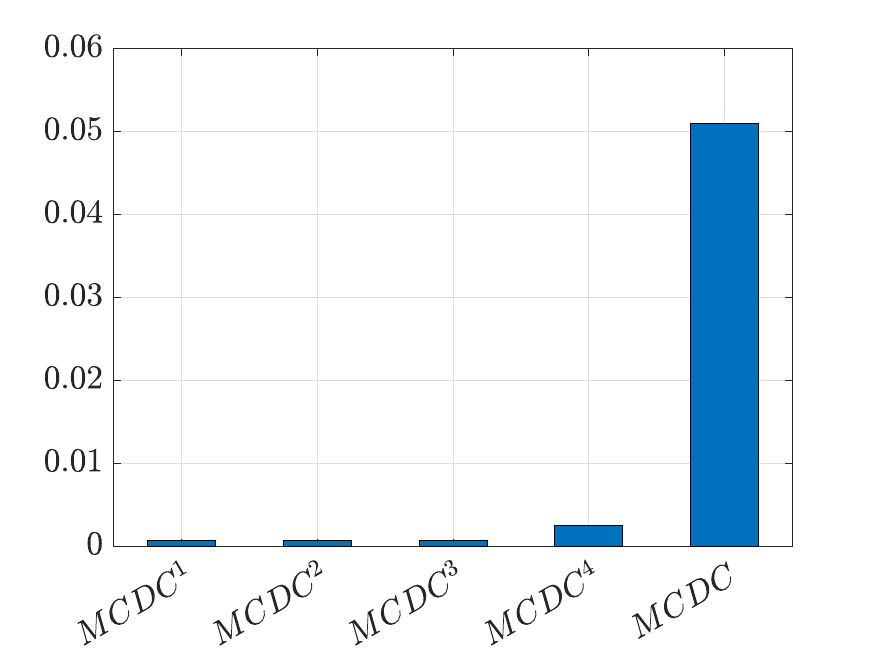}
		\caption{ARI on Car.}
		\label{fig:sub1.1}
	\end{subfigure}
	\hfill
	\begin{subfigure}{0.49\columnwidth}
		\includegraphics[width=1.1\linewidth]{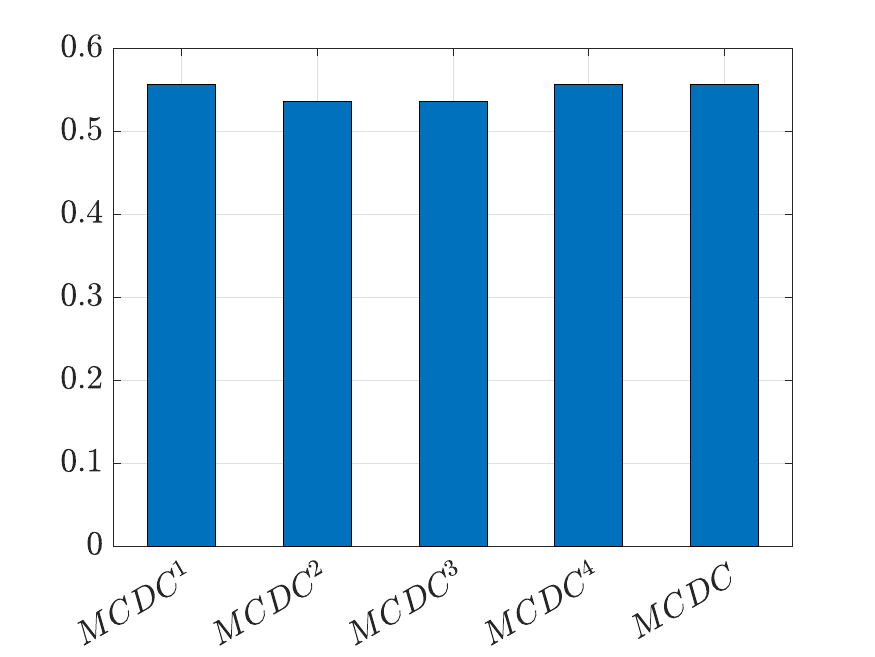}
		\caption{ARI on Con.}
		\label{fig:sub1.2}
	\end{subfigure}
	\par\medskip
	\begin{subfigure}{0.49\columnwidth}
		\includegraphics[width=1.1\linewidth]{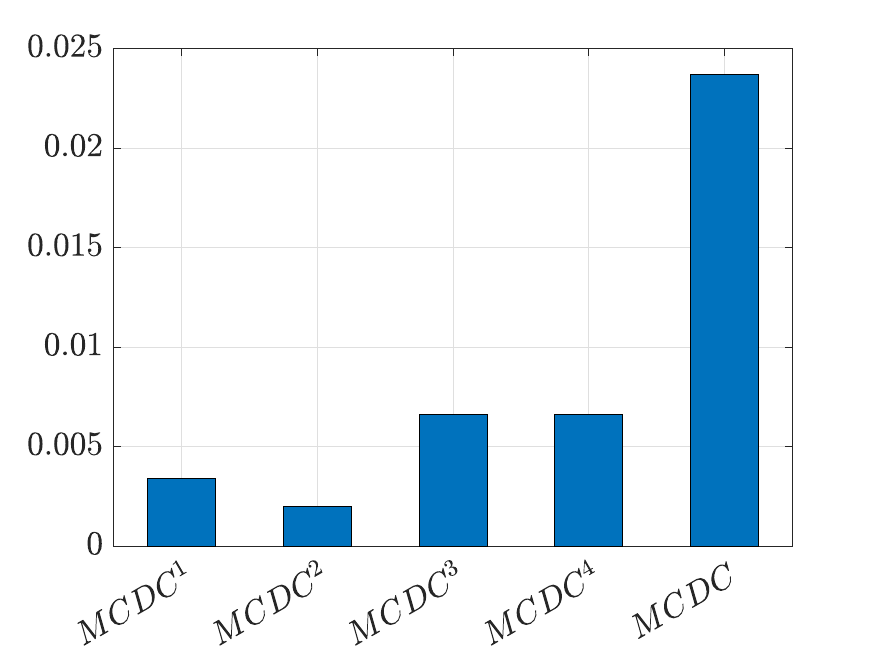}
		\caption{ARI on Che.}
		\label{fig:sub1.3}
	\end{subfigure}
	\hfill
	\begin{subfigure}{0.49\columnwidth}
		\includegraphics[width=1.1\linewidth]{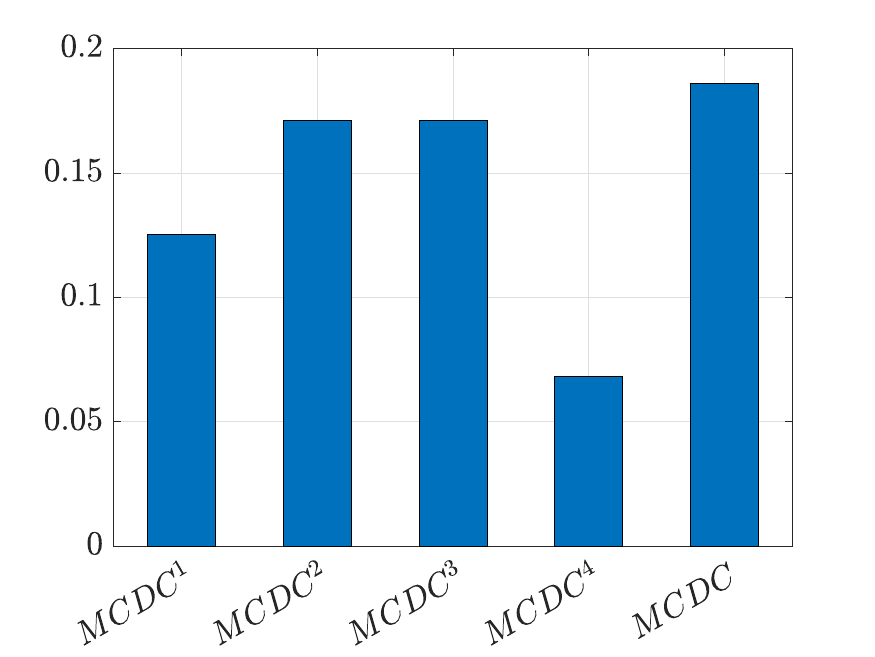}
		\caption{ARI on Mus.}
		\label{fig:sub1.4}
	\end{subfigure}
	\par\medskip
	\begin{subfigure}{0.49\columnwidth}
		\includegraphics[width=1.1\linewidth]{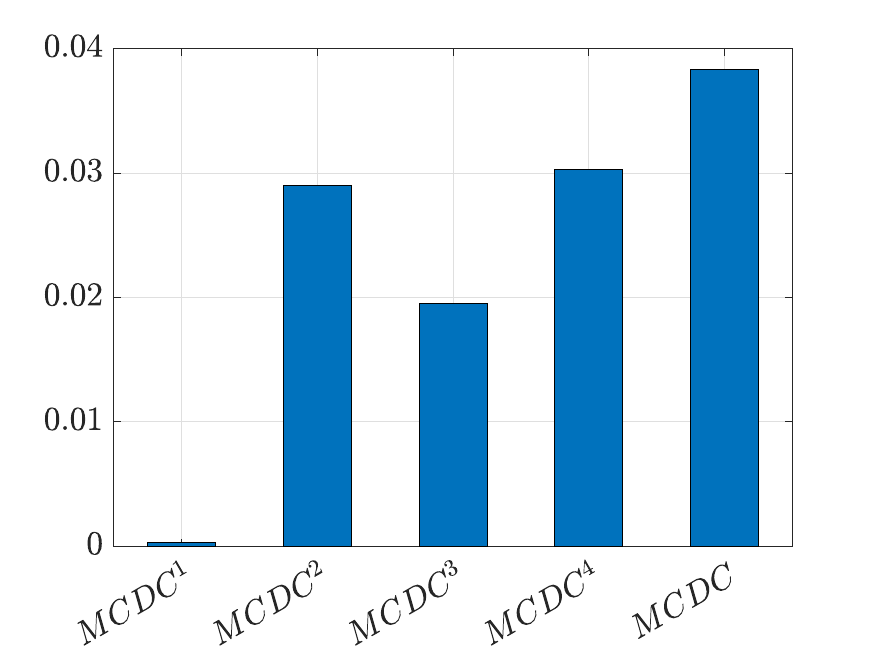}
		\caption{ARI on Tic.}
		\label{fig:sub1.5}
	\end{subfigure}
	\hfill
	\begin{subfigure}{0.49\columnwidth}
		\includegraphics[width=1.1\linewidth]{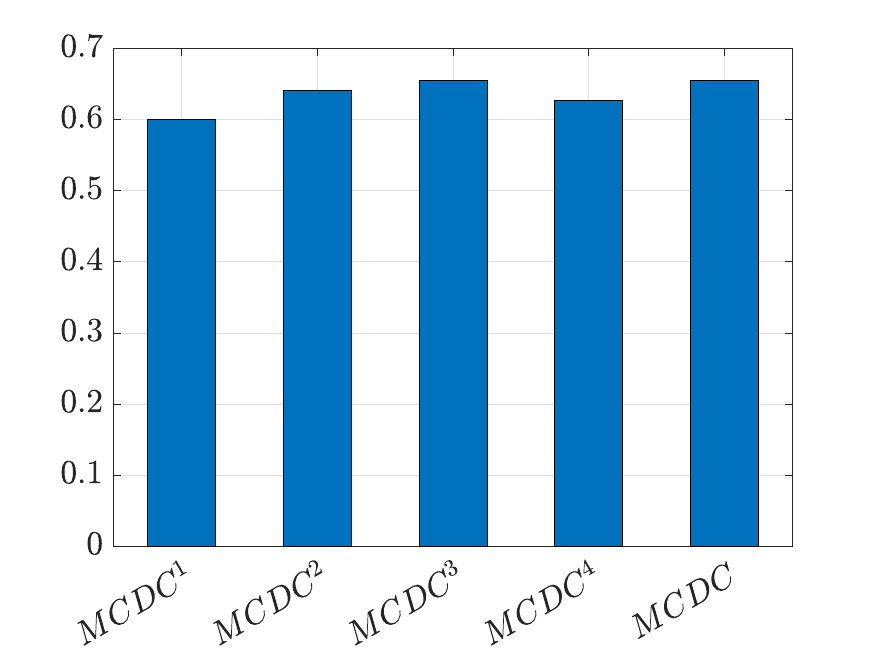}
		\caption{ARI on Vot.}
		\label{fig:sub1.6}
	\end{subfigure}
	\par\medskip
	\begin{subfigure}{0.49\columnwidth}
		\includegraphics[width=1.1\linewidth]{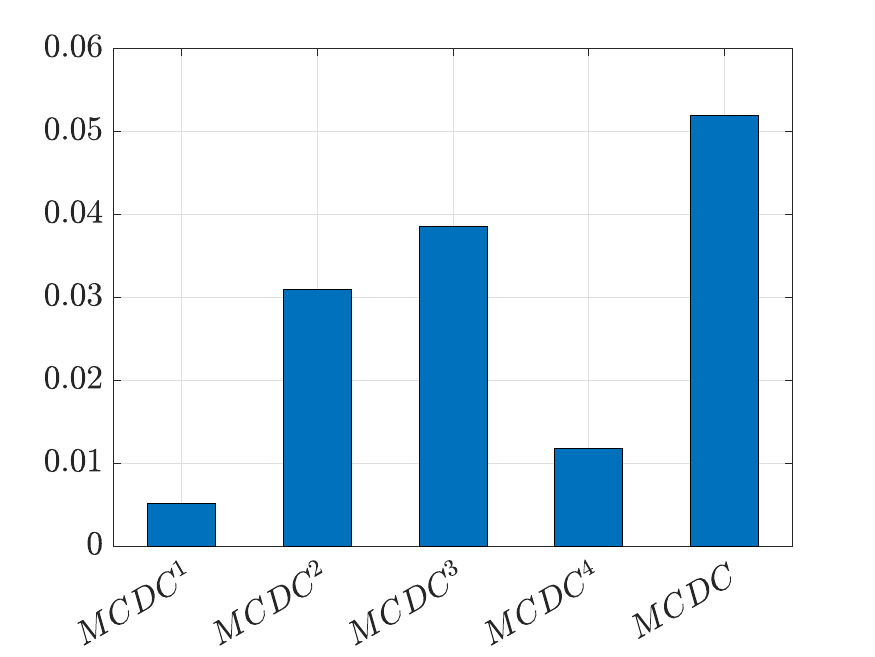}
		\caption{ARI on Bal.}
		\label{fig:sub1.7}
	\end{subfigure}
	\hfill
	\begin{subfigure}{0.49\columnwidth}
		\includegraphics[width=1.1\linewidth]{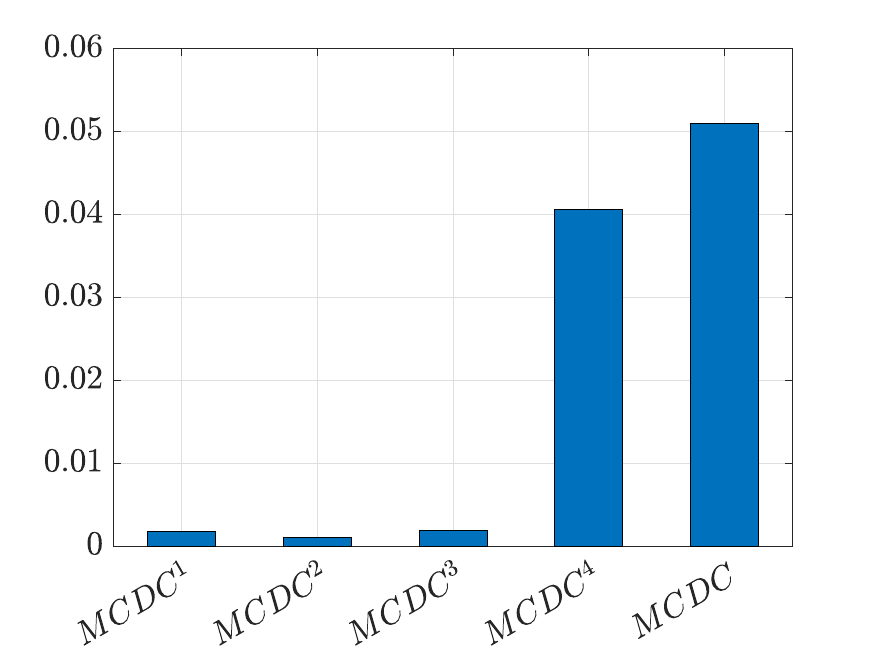}
		\caption{ARI on Nur.}
		\label{fig:sub1.8}
	\end{subfigure}
	\caption{Comparison of MCDC and its four ablated versions, i.e., MCDC$^4$, MCDC$^3$, MCDC$^2$, and MCDC$^1$, which are obtained by removing the weighting mechanism of CAME, the whole CAME, multi-granular learning mechanism of MGCPL, and the whole MGCPL from MCDC in turn.}
	\label{fig:1}
\end{figure}

\begin{figure}[!t]
	\centering
	\begin{subfigure}{0.49\columnwidth}
		\includegraphics[width=1.1\linewidth]{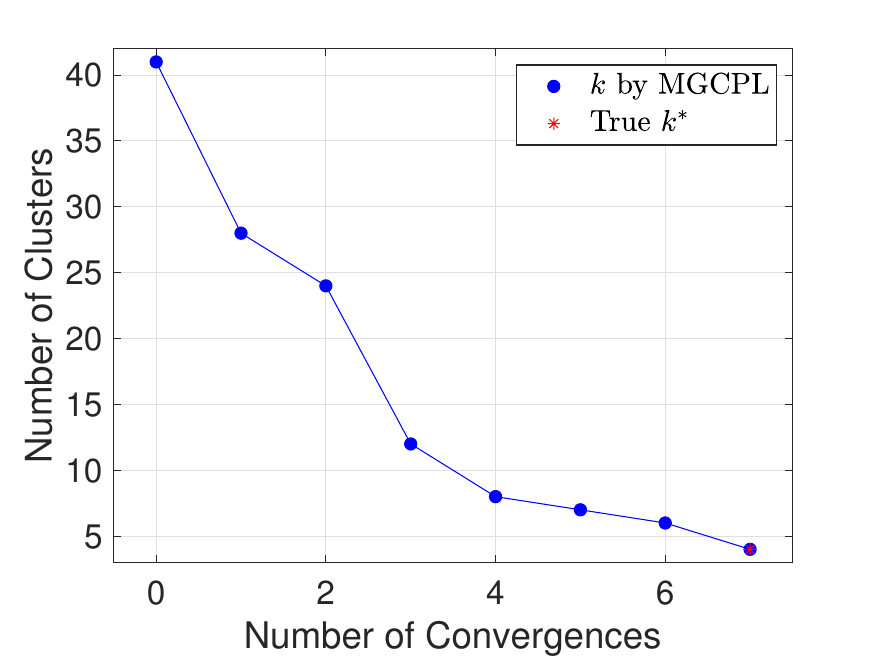}
		\caption{$k$s learned for Car.}
		\label{fig:sub2.1}
	\end{subfigure}
	\hfill
	\begin{subfigure}{0.49\columnwidth}
		\includegraphics[width=1.1\linewidth]{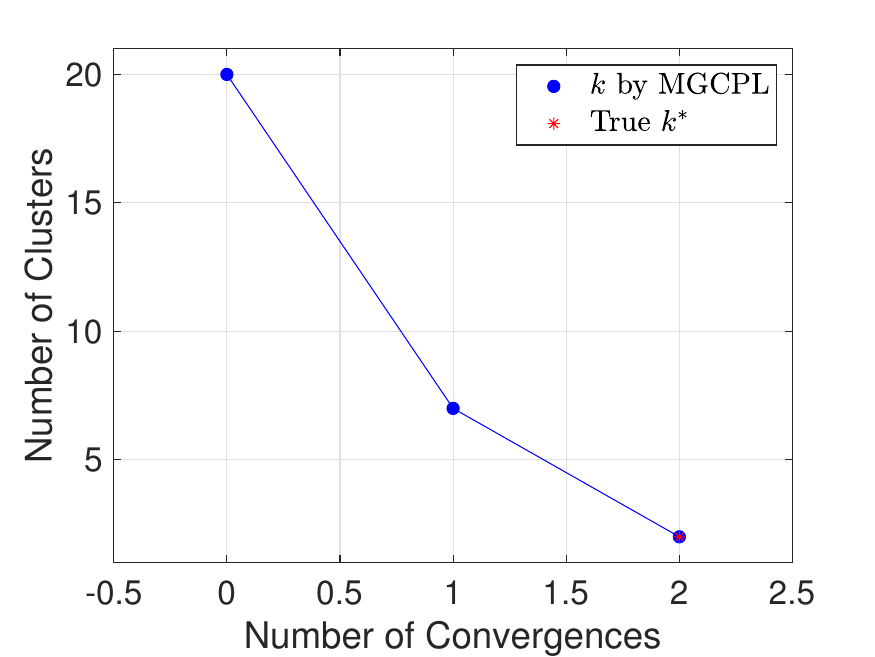}
		\caption{$k$s learned for Con.}
		\label{fig:sub2.2}
	\end{subfigure}
	\par\medskip
	\begin{subfigure}{0.49\columnwidth}
		\includegraphics[width=1.1\linewidth]{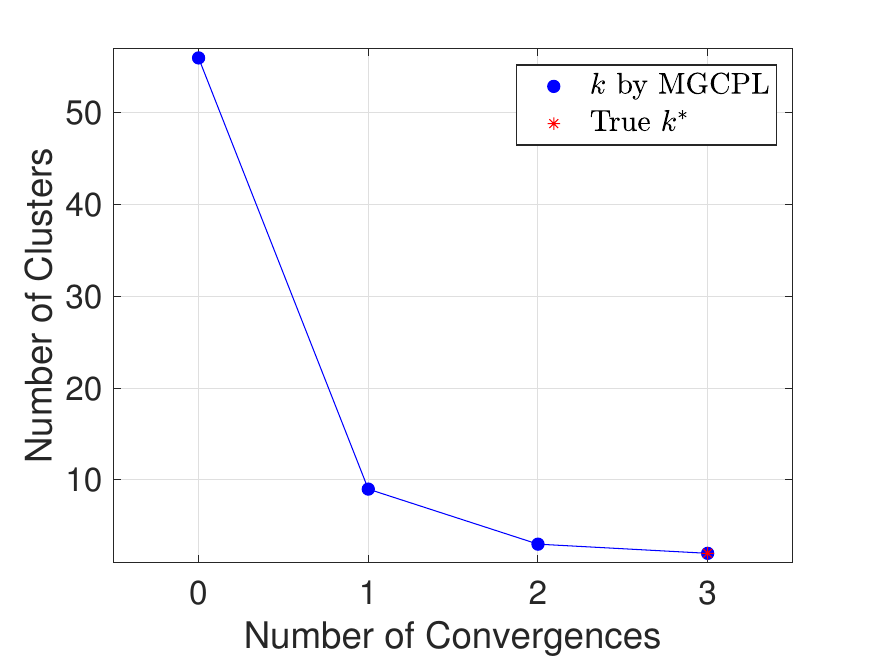}
		\caption{$k$s learned for Che}
		\label{fig:sub2.3}
	\end{subfigure}
	\hfill
	\begin{subfigure}{0.49\columnwidth}
		\includegraphics[width=1.1\linewidth]{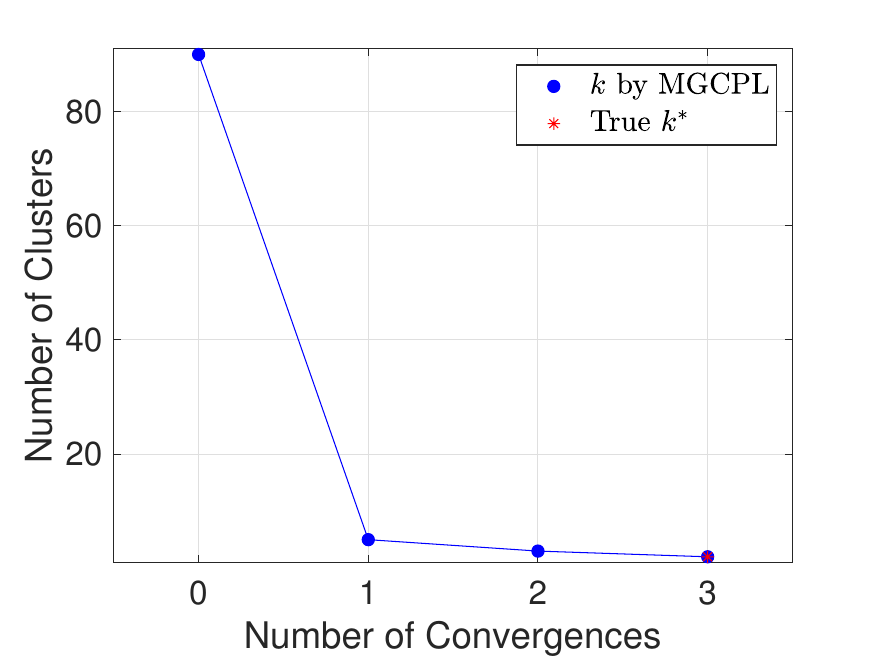}
		\caption{$k$s learned for Mus.}
		\label{fig:sub2.4}
	\end{subfigure}
	\par\medskip
	\begin{subfigure}{0.49\columnwidth}
		\includegraphics[width=1.1\linewidth]{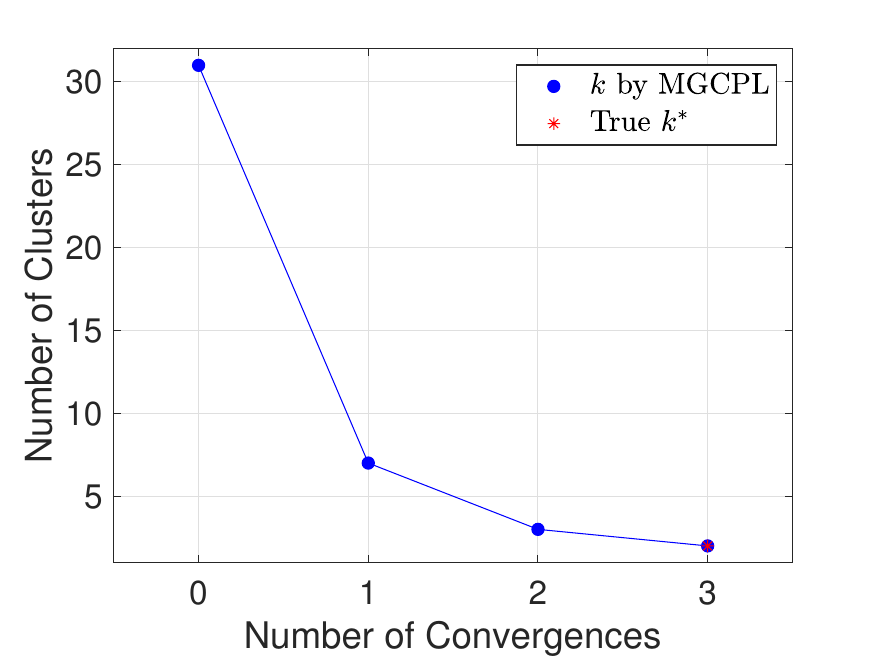}
		\caption{$k$s learned for Tic.}
		\label{fig:sub2.5}
	\end{subfigure}
	\hfill
	\begin{subfigure}{0.49\columnwidth}
		\includegraphics[width=1.1\linewidth]{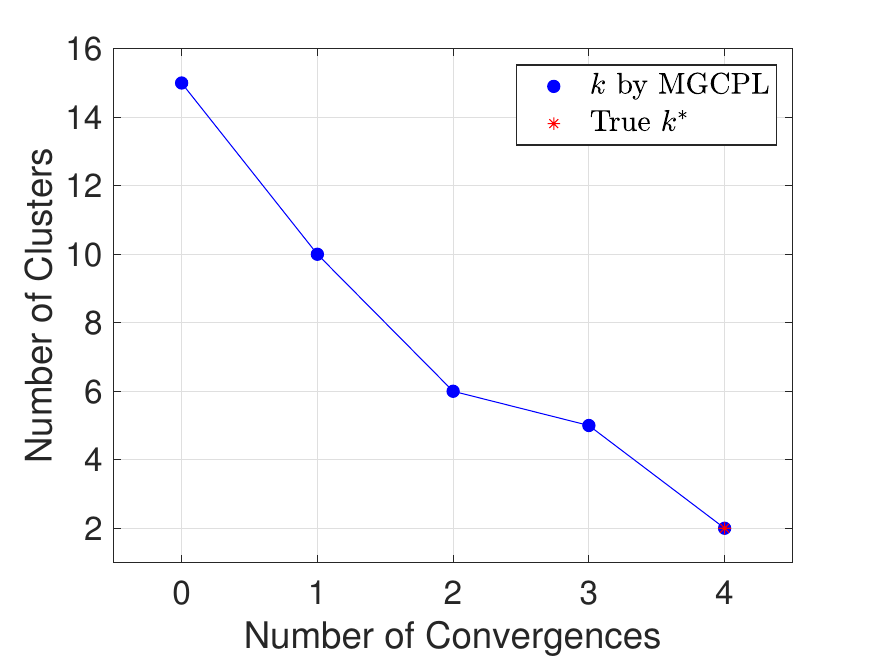}
		\caption{$k$s learned for Vot.}
		\label{fig:sub2.6}
	\end{subfigure}
	\par\medskip
	\begin{subfigure}{0.49\columnwidth}
		\includegraphics[width=1.1\linewidth]{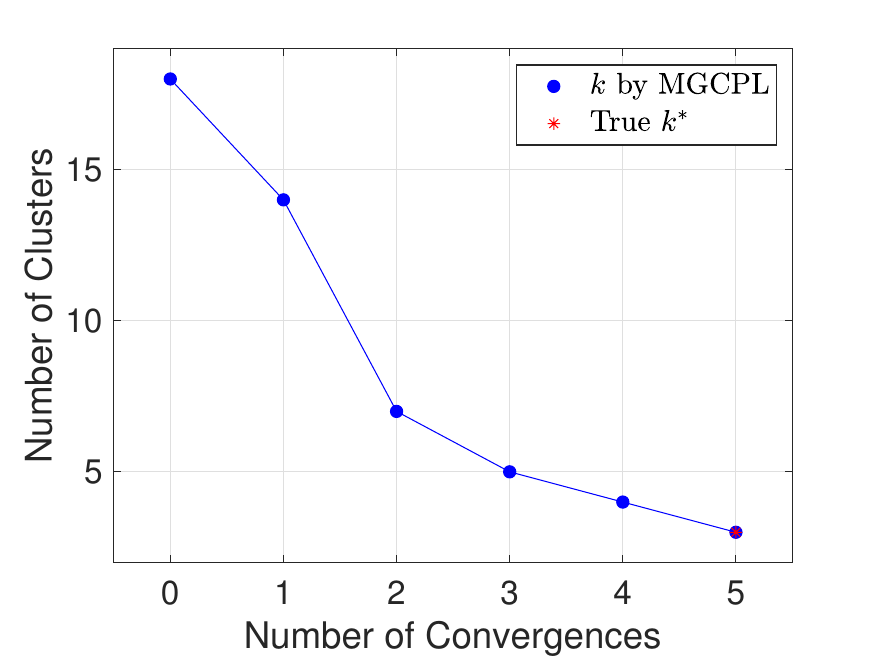}
		\caption{$k$s learned for Bal.}
		\label{fig:sub2.7}
	\end{subfigure}
	\hfill
	\begin{subfigure}{0.49\columnwidth}
		\includegraphics[width=1.1\linewidth]{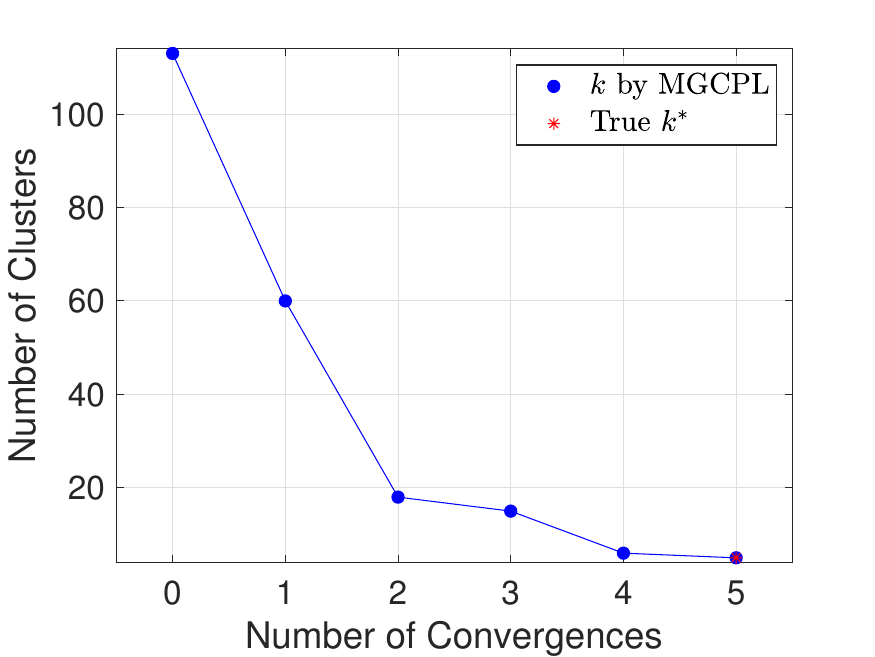}
		\caption{$k$s learned for Nur.}
		\label{fig:sub2.8}
	\end{subfigure}
	\caption{Different numbers of clusters learned by MGCPL. Blue dots indicate the number of clusters when MGCPL temporarily converges under the current cluster granularity. Red stars indicate the true number of clusters $k^*$.}
	\label{fig:2}
\end{figure}

To further verify the effectiveness of different main components of the proposed MCDC method, we ablate it into the following four versions: 1) MCDC$^4$ is the version that replaces the feature weighting mechanism (described by Eqs.~(\ref{eq:theta})-(\ref{eq:I}) in Section~\ref{sct:came}) of CAME with fixed identical weights, 2) MCDC$^3$ is obtained by removing the whole CAME module from MCDC and use the $k_\sigma$ learned by MGCPL for clustering, 3) MCDC$^2$ is the version obtained by replacing MGCPL of MCDC$^3$ with the conventional competitive learning with $k^*+2$ as the initialization described in Section~\ref{sct:cpl}, and 4) MCDC$^1$ is the version formed by further removing the competitive learning mechanism from MCDC$^2$ and only adopt the object-cluster distance described in Section~\ref{sct:ocd}. Since the version of MCDC$^1$ that replaces object-cluster distance with the conventional Hamming distance metric is equivalent to $k$-modes, which has been compared in Table~\ref{tbl:3}, We do not further ablate MCDC$^1$ to avoid duplicated results.

It can be seen from the results shown in Fig. \ref{fig:1} that the ARI performance of MCDC, MCDC$^4$, MCDC$^3$, MCDC$^2$, MCDC$^1$ sequentially decreases in general, which intuitively illustrate the effectiveness of all the proposed main technical components of MCDC.

More specifically, it can be observed that MCDC always outperforms MCDC$^4$. This indicates that the feature weighting mechanism in CAME (i.e., Eqs.~(\ref{eq:theta})-(\ref{eq:I}) in Section~\ref{sct:came}) is effect in learning the importance of features in the embeddings output by CAME, and also illustrates that the encoding strategy of CAME is effective in fusing the multi-granular information provided by MGCPL.

It can also be observed that MCDC$^4$ performs not worse than MCDC$^3$ on five data sets, but outperformed by MCDC$^3$ on three data sets, i.e., Mus., Vot., and Bal. This is because identical weights of MCDC$^4$ cannot appropriately reflect the importance of the encoded features of these three data sets, which again highlights the necessity of weights learning.

The effectiveness of the proposed MGCPL is illustrated by the fact that MCDC$^3$ outperforms MCDC$^2$ on almost all the data sets. The reason is that MGCPL learns the cluster distributions from different $k$s. Although the result of MCDC$^3$ is obtained at the final $k_\sigma$, the previous $k_{\sigma-1}$ learned by MGCPL provides a more reasonable initialization for the last round learning compared to the initialized $k$ of MCDC$^2$ where $k=k^*+2$ .

By comparing MCDC$^2$ and MCDC$^1$, it can be found that MCDC$^2$ has no significant advantage over MCDC$^1$. The reason is that MCDC$^1$ requires $k^*$ to be given in advance for clustering, while MCDC$^2$ automatically learns to find $k^*$. In other words, although MCDC$^2$ adopting a competitive learning mechanism is more powerful, its advantage is obscured because $k^*$ is unfairly leaked to MCDC$^1$.

\subsection{Learning Process Evaluation}

Numbers of clusters, i.e., $\kappa=\{k_1,k_2,...,k_\sigma\}$, learned by MGCPL are demonstrated in Fig.~\ref{fig:2} where blue dots indicate the number of clusters when MGCPL temporarily converges under the current cluster granularity, and red stars indicate the true number of clusters $k^*$ as shown in Table~\ref{tbl:2}. Please note that the number of clusters corresponding to ``0'' on the x-axis indicates the initialized $k$.

It can be observed that MGCPL converges in stages during its learning, which reflects that MGCPL can automatically learn clusters with different granularities. It can also be observed that almost all the final $k_\sigma$ learned by MGCPL equal to the true number of clusters $k^*$, which indicates that MGCPL is competent in searching for the optimal number of clusters $k^*$ without prior clustering knowledge.

\subsection{Computational Efficiency Evaluation}

\begin{figure}[!t]
	\centering
	\begin{subfigure}{\columnwidth}
		\centering
		\includegraphics[width=1.08\linewidth]{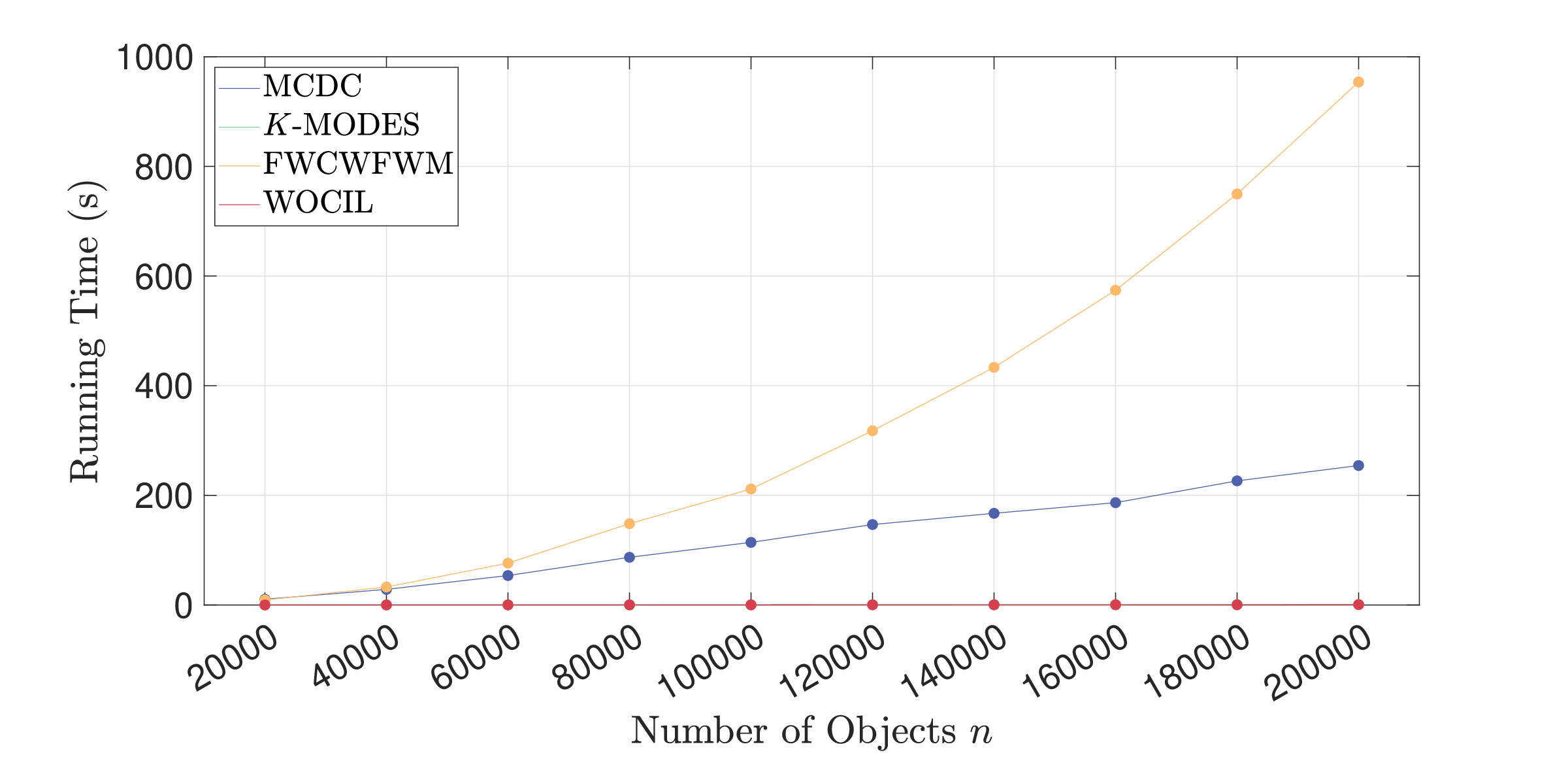}
		\caption{Time on Syn\_n w.r.t. $n$}
	\end{subfigure}
	\begin{subfigure}{\columnwidth}
		\centering
		\includegraphics[width=1.08\linewidth]{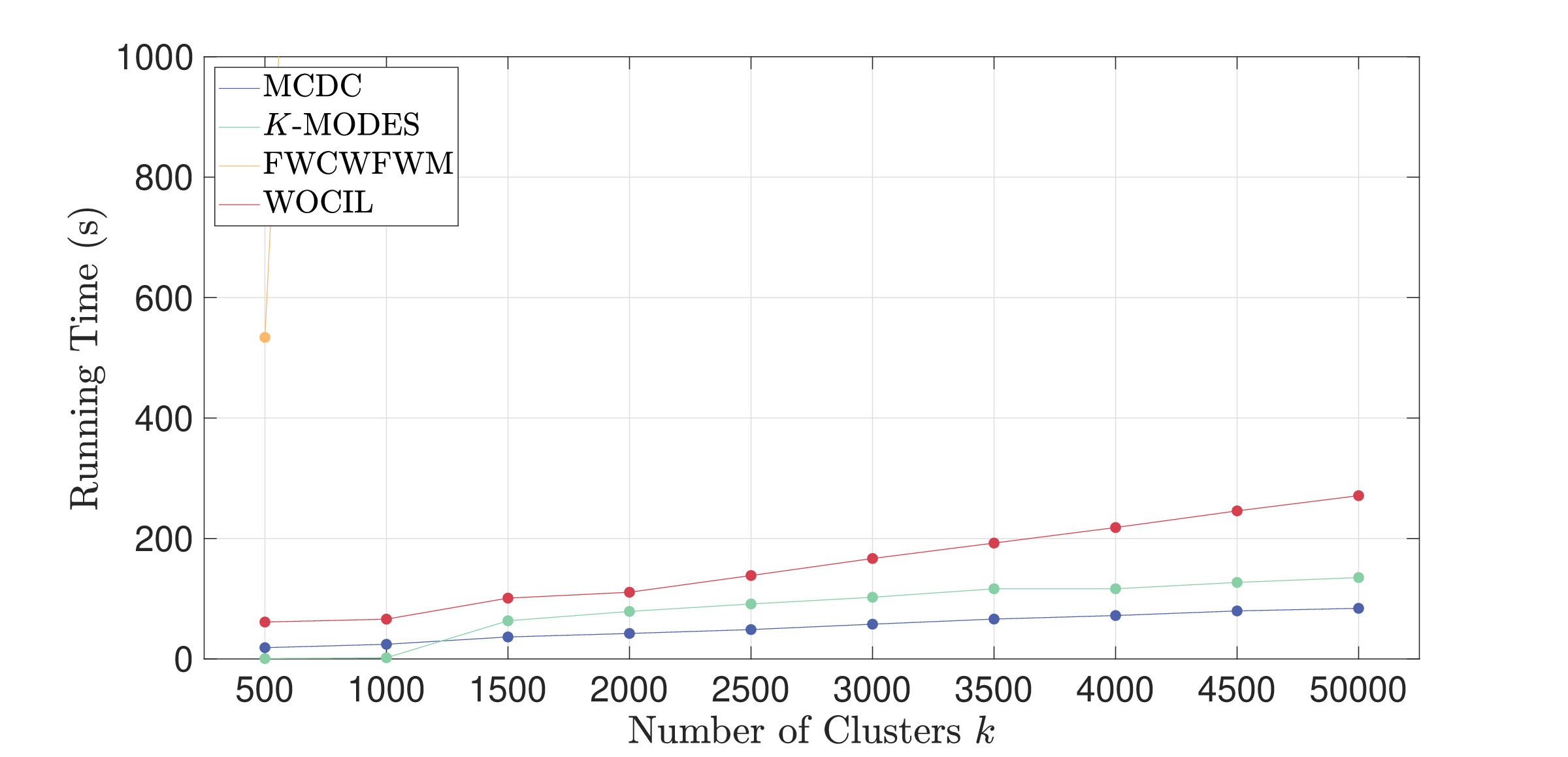}
		\caption{Time on Syn\_n w.r.t. $k$}
	\end{subfigure}
	\begin{subfigure}{\columnwidth}
		\centering
		\includegraphics[width=1.08\linewidth]{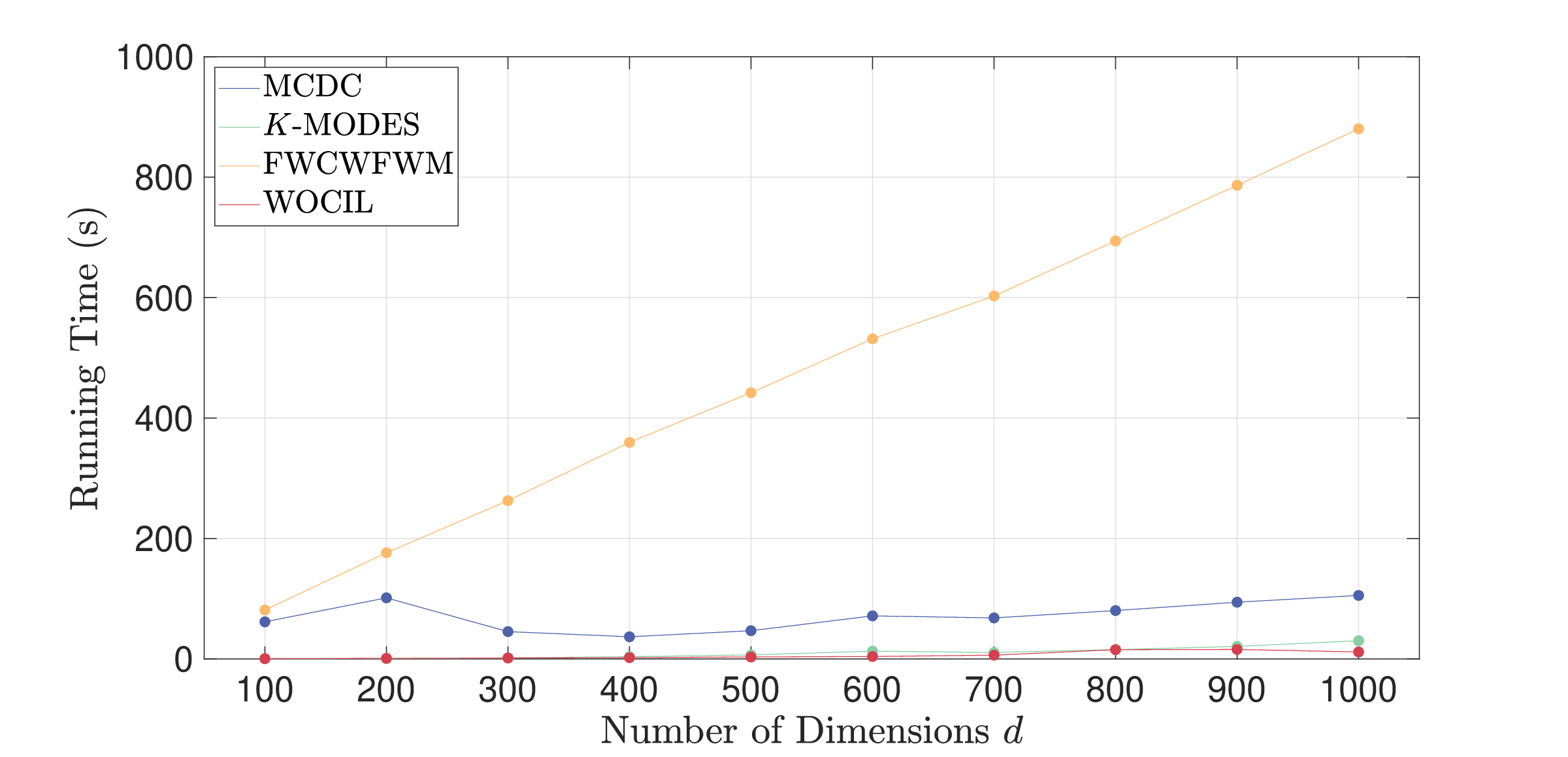}
		\caption{Time on Syn\_d w.r.t. $d$}
	\end{subfigure}
	\caption{Execution time of different methods on (a) Syn\_n, (b) Syn\_n, and (c) Syn\_d with increasing $n$, $k$, and $d$, respectively.}
	\label{fig:3}
\end{figure}

The execution time of MCDC on three synthetic data sets is shown in Fig.~\ref{fig:3}. We implement several representative counterparts on each synthetic data set with different $n$s, $k$s, and $d$s to verify the time complexity of MCDC. Note that $k$ in this experiment is the number of sought clusters $k$ in Algorithm~\ref{alg:2}. The execution time is averaged on ten runs of the corresponding method.

Intuitively, the execution time of MCDC increases linearly with the increasing of data size $n$ and feature scale $d$, which confirms our analysis at the end of Section~\ref{sct:came} that MCDC is with linear time complexity w.r.t. $n$ and $d$. Moreover, it can also be observed that MCDC has linear time complexity w.r.t. $k$, which indicates that MCDC can be easily applied to different clustering tasks of customized $k$. In general, MCDC is scalable to large-scale categorical data.

\section{Concluding Remarks}
This paper proposes a new method called MCDC for cluster analysis of categorical data. MCDC is composed of MGCPL for learning nested multi-granular cluster distribution, and CAME for aggregating the learned nested distribution to obtain partitional clustering results. As the learning process of MGCPL is fully automatic and highly interpretable, the complex cluster distribution of categorical data can be intuitively revealed. Accordingly, CAME encodes the multi-granular distribution learned by MGCPL to obtain informative embeddings of data objects for clustering. Since the two main components of MCDC, i.e., MGCPL and CAME, are both with linear time complexity, MCDC is scalable to large-scale categorical data. Extensive experiments illustrate the superiority of MCDC in terms of clustering accuracy, robustness to data sets in different fields, and computational efficiency.

Some limitations of this work include: 1) the proposed method has not been extended to more complex heterogeneous feature data and multi-modal data and 2) we assumed that the data is static and has not yet considered more complex dynamically distributed data clustering. Building on this research, some promising future research orientations are: 1) applying MGCPL to discover implicit cluster distributions in different fields, 2) extending the whole MCDC to process streaming and dynamic data, and 3) leveraging the advantages of MGCPL to active learning for reducing the workload of human experts in manually labeling large-scale categorical data sets.

\section*{Acknowledgements}

This work was supported in part by the National Natural Science Foundation of China (NSFC) under grants: 62102097, 62374047, and 62174038, the NSFC/Research Grants Council (RGC) Joint Research Scheme under the grant N\_HKBU214/21, the Natural Science Foundation of Guangdong Province under grants: 2023A1515012855 and 2022A1515011592, the Guangdong Provincial Key Laboratory under grant 2023B1212060076, the General Research Fund of RGC under grants: 12201321, 12202622, and 12201323, the RGC Senior Research Fellow Scheme under grant SRFS2324-2S02, and the Science and Technology Program of Guangzhou under grant 202201010548.

\normalem
\bibliographystyle{IEEEtran}
\bibliography{References}

\end{document}